\documentclass{article}

\usepackage[preprint]{neurips_2026}

\usepackage[utf8]{inputenc} 
\usepackage[T1]{fontenc}    
\usepackage{hyperref}       
\usepackage{url}            
\usepackage{xurl}           
\usepackage{booktabs}       
\usepackage{amsfonts}       
\usepackage{nicefrac}       
\usepackage{microtype}      
\usepackage{xcolor}         
\usepackage{colortbl}       

\usepackage{graphicx}
\usepackage{amsmath}
\usepackage{subcaption}
\usepackage{xspace}
\usepackage{tcolorbox}
\usepackage{adjustbox}
\usepackage{array}
\usepackage{algorithm}
\usepackage{algorithmic}
\usepackage{placeins}
\usepackage{tabularx}
\usepackage{needspace}      

\newcommand{\methodName}{DISCO\xspace}

\definecolor{second}{RGB}{235, 247, 255}
\definecolor{best}{RGB}{210, 235, 255}
\newcommand{\best}[1]{\cellcolor{best}\textbf{#1}}
\newcommand{\second}[1]{\cellcolor{second}\textbf{#1}}

\newif\ifshowedits
\showeditstrue

\definecolor{editmajor}{RGB}{190, 0, 0}      
\definecolor{editmoderate}{RGB}{200, 95, 0}  
\definecolor{editminor}{RGB}{0, 70, 190}     
\definecolor{editcite}{RGB}{0, 130, 60}      
\definecolor{edittodo}{RGB}{255, 240, 150}   

\ifshowedits
  \newcommand{\major}[1]{{\color{editmajor}#1}}

  \newcommand{\todo}[1]{\colorbox{edittodo}{\parbox{\dimexpr\linewidth-2\fboxsep}{\small\textbf{TODO:} #1}}}
  \newcommand{\todoinline}[1]{\colorbox{edittodo}{\small\textbf{TODO:} #1}}
  \newcommand{\todocell}[1]{\colorbox{edittodo}{#1}}
\else
  \newcommand{\major}[1]{#1}

  \newcommand{\todo}[1]{}
  \newcommand{\todoinline}[1]{}
  \newcommand{\todocell}[1]{}
\fi

\usepackage{tikz}
\definecolor{unibernred}{RGB}{238,16,30}
\newlength{\unibernLogoHeight}

\newcommand{\unibernWordLine}[3]{%
\node[
anchor=base west,
inner sep=0pt,
text=unibernred,
font=\fontfamily{phv}\bfseries\fontsize{120}{120}\selectfont,
xscale=1.09
] at (#1,#2) {#3};
}

\newcommand{\unibernTikZLogo}[1][0.75cm]{%
\begingroup
\setlength{\unibernLogoHeight}{#1}%
\pgfmathsetmacro{\unibernScale}{\the\unibernLogoHeight/335pt}%
\begin{tikzpicture}[x=1pt,y=1pt,scale=\unibernScale,transform shape]
\path[use as bounding box] (0,0) rectangle (1012,335);

\node[
anchor=base west,
inner sep=0pt,
text=black,
font=\fontfamily{ptm}\bfseries\itshape\fontsize{240}{240}\selectfont,
xscale=1.02,
yscale=0.98
] at (34,61) {u};

\node[
anchor=base west,
inner sep=0pt,
text=black,
font=\fontfamily{ptm}\itshape\fontsize{122}{122}\selectfont,
xscale=0.86,
yscale=1.00
] at (190,172) {b};

\unibernWordLine{305}{130}{UNIVERSITY}
\unibernWordLine{305}{10}{OF BERN}
\end{tikzpicture}%
\endgroup
}

\newcommand{\unibernTikZLogoHor}[1][0.75cm]{%
\begingroup
\setlength{\unibernLogoHeight}{#1}%
\pgfmathsetmacro{\unibernScale}{\the\unibernLogoHeight/335pt}%
\begin{tikzpicture}[x=1pt,y=1pt,scale=\unibernScale,transform shape]
\path[use as bounding box] (0,0) rectangle (1900,335);

\node[
anchor=base west,
inner sep=0pt,
text=black,
font=\fontfamily{ptm}\bfseries\itshape\fontsize{240}{240}\selectfont,
xscale=1.02,
yscale=0.98
] at (34,61) {u};

\node[
anchor=base west,
inner sep=0pt,
text=black,
font=\fontfamily{ptm}\itshape\fontsize{122}{122}\selectfont,
xscale=0.86,
yscale=1.00
] at (190,170) {b};

\node[
anchor=base west,
inner sep=0pt,
text=unibernred,
font=\fontfamily{phv}\bfseries\fontsize{165}{165}\selectfont,
xscale=1.02
] at (280,55) {University of Bern};
\end{tikzpicture}%
\endgroup
}

\title{\hyphenpenalty=10000 \exhyphenpenalty=10000 Emergent Specialization in Populations of Self-Supervised Collaborative Vision Experts Without a Shared Gate or Cross-Agent Gradients}

\author{%
  Aram Davtyan \\
  \texttt{aram.davtyan@unibe.ch}
  \And
  Pablo Acuaviva \\
  \texttt{pablo.acuavivahuertos@unibe.ch}
  \And
  Sebastian Stapf \\
  \texttt{sebastian.stapf@unibe.ch}
  \And
  Paolo Favaro \\
  \texttt{paolo.favaro@unibe.ch}
  \AND
  \begin{tabular}{@{}c@{}}
  \unibernTikZLogoHor[0.7cm]\\[6pt]
  Computer Vision Group, Institute of Computer Science\\
  University of Bern, Switzerland
  \end{tabular}
}

\begin{document}

\maketitle

\begin{abstract}
Can a population of neural networks develop a useful division of labor without a shared gate or gradients between agents? We study a setting where each network has its own weights, trains independently on the same heterogeneous data, and can ask another agent for help through a forward pass. Unlike mixtures of experts, where a jointly trained gate assigns inputs to experts, specialization here must emerge without central control. 
We test this in a small scale proxy for predictive visual pretraining. Initially identical agents are finetuned on an unlabeled mixture of six visual domains using masked prediction of frozen DINOv3 features. We measure \emph{specialization} by asking whether the best agent for an input aligns with its latent domain, and \emph{utilization} by asking whether responsibility is distributed across agents. We progressively remove central control, ending with \methodName (DIStributed COllaboration) where each agent locally selects a helper, reads its internal state through a gradient free channel, and rewards its router only for the improvement that help provides.
Specialization emerges and is useful. Randomly routed populations underperform a single generalist, while semantically routed populations outperform it, showing that specialization rather than population size drives the gain. Specialization persists without a central router, and gradient free communication lets nonexperts exploit emergent expertise. In \methodName, a random agent helped by the expert matches the solo generalist, while experts surpass it, including on data outside the specialization mixture. Local routers select the emergent expert for 98\% of inputs. These effects persist across population size, model capacity, data imbalance, and finetuning seeds, providing measurable evidence for the dynamics needed by decentralized predictive pretraining.
\end{abstract}

\section{Introduction}\label{sec:introduction}

Visual foundation models are mostly trained as single monolithic networks: one set of parameters processes every image, regardless of its domain, content, or source~\citep{dosovitskiy2020image,oquab2024dinov2,simeoni2025dinov3}, and the same holds for video world models trained on increasingly heterogeneous data~\citep{bruce2024genie,nvidia2025cosmos,assran2025vjepa2}. This works well, but leaves a basic question open: is a single network the right way to allocate capacity over heterogeneous data?

The alternative is a division of labor. Machine learning has explored it mainly through mixtures of experts (MoE)~\citep{jacobs1991adaptive,shazeer2017outrageously,fedus2022switch}, including sparse vision transformers~\citep{riquelme2021scaling,puigcerver2024sparse} and expert denoisers in diffusion models~\citep{balaji2022ediffi,feng2023ernievilg2,fei2024ditmoe}. However, specialization in MoE is typically assumed rather than measured, and when it is measured, experts often organize around token position, noise level, or a small set of always-active experts rather than around semantic content~\citep{fei2024ditmoe,wei2026routing,wang2026illusion,huang2026sdmoe}. Moreover, MoE specialization is orchestrated: one gate, trained jointly with all experts in one computational graph, decides who processes what. This paper asks whether a division of labor also emerges among \emph{separate} networks that route locally, without a shared gate and without gradients flowing between them, and whether it is of any use once it does, a question that has not been tested in a controlled way.

\paragraph{Why this setting matters.} This is a candidate regime for training populations of models at scale. Without a shared gate, no global component lies on the path of every sample. Without cross-agent gradients, each agent is an independent optimization problem that can in principle run on separate hardware, be updated asynchronously, and be added to or removed from the population. Populations of separately trained models are already used to scale beyond one network, as branched and merged experts~\citep{li2022branch,gururangan2023scaling,sukhbaatar2024branch}, decentralized paths~\citep{douillard2024dipaco}, or domain ensembles~\citep{ersoy2025hdee}. All of this relies on an assumption that has not been tested in isolation: that a population of models, on its own, divides labor along the structure of its data, and that what one member learns can be used by the others. If it holds, decentralized populations become a candidate route to scaling on heterogeneous data; if not, central orchestration is necessary rather than optional.

\paragraph{A controlled proxy for predictive pretraining.} Self-supervised pretraining is dominated by the prediction of missing or upcoming content: the next token in language models~\citep{brown2020language}, the next frame in video world models~\citep{bruce2024genie,nvidia2025cosmos}, and masked or future features in joint-embedding predictive architectures~\citep{wei2022maskfeat,assran2023ijepa,bardes2024revisiting} and latent world models~\citep{dinowm,karypidis2025dino}. Whether populations of such predictors develop the dynamics studied here should be verified at small scale first. We therefore use the simplest task that preserves the operation of predictive pretraining: agents reconstruct 70\% masked patches in the frozen feature space of DINOv3~\citep{simeoni2025dinov3}. The data are an unlabeled mixture of six visual domains outside the pretraining distribution, whose identity is the hidden factor that lets us measure specialization. Training a population takes hours, which makes a controlled study over regimes, population sizes, and capacities possible (Section~\ref{sec:setting}).

\paragraph{Approach and questions.} $K$ agents start from the same pretrained weights and are fine-tuned with one shared self-supervised objective on the mixture, with no labels, roles, or central controller. We approach the target setting step by step along a ladder of training regimes with decreasing central control (Table~\ref{tab:roadmap}) and ask: \emph{(Q1)} Does a division of labor aligned with the latent domains improve on a single generalist, and is the improvement due to specialization rather than to having more models? \emph{(Q2)} Does specialization still emerge when the central router is removed and every agent routes locally? \emph{(Q3)} Can the expertise that emerges in one agent be used by another through a channel that carries no gradients? Three elements make the questions answerable. \emph{(a) Metrics}: specialization is the normalized mutual information between the identity of the best agent for an input and the latent domain, and utilization the normalized entropy of that assignment; together they separate a division of labor from collapse and from load balancing alone. \emph{(b) A testbed} with known latent structure in which all regimes share agents, initialization, and data budget, so that differences are attributable to routing and communication alone. \emph{(c) The ladder}: a solo model; populations with random, privileged K-means, and learned central routing; \emph{distributed delegation}, where each agent owns a router; and \emph{\methodName} (DIStributed COllaboration), the target setting, where a requester agent selects a helper, reads the helper's internal state through a stop-gradient channel, and rewards its router only for the local improvement the help produces.

\paragraph{Findings.} Specialization emerges in the target setting, and it is useful. A randomly routed population of the same size and budget is worse than the solo model while semantically routed populations are better, so specialization and not population size is what helps (Section~\ref{sec:res_central}). Specialization survives the removal of the central router (Section~\ref{sec:res_delegation}). The gradient-free exchange makes the emergent expertise usable: \methodName is the only regime in which a helped non-expert matches the solo model, it has the best worst-agent performance, and the effect persists on the ImageNet100 validation set, which the agents saw during pretraining but not during specialization (Section~\ref{sec:res_disco}). Local routers trained only on their own improvement select the emergent expert for 98\% of inputs (Section~\ref{sec:res_routing}), and the onset of specialization depends on population size and capacity in a predictable way (Section~\ref{sec:res_scaling}).

\section{Related Work}\label{sec:prior_work}

\paragraph{Expert specialization in vision and generative models.}
Mixtures of experts route inputs to gated sub-networks~\citep{jacobs1991adaptive,shazeer2017outrageously,fedus2022switch,zhou2022expertchoice}; in vision they are used in ViTs~\citep{riquelme2021scaling,puigcerver2024sparse,han2025vimoe}, masked autoencoders~\citep{liu2024moce}, and diffusion transformers~(\citealp{fei2024ditmoe}; \citealp{shi2025diffmoe}), and diffusion models have been split into expert denoisers by noise level~\citep{balaji2022ediffi,feng2023ernievilg2}. In practice, gated specialization is fragile: representations collapse around expert centroids~\citep{chi2022representation}; diffusion experts specialize by position and timestep rather than content~\citep{fei2024ditmoe}, motivating routing guidance~\citep{wei2026routing}, expert competition~\citep{zheng2025gatepro}, or spectral decomposition~\citep{huang2026sdmoe}; and MoE language models exhibit a domain-invariant ``standing committee''~\citep{wang2026illusion}. All of these study experts inside one jointly optimized network with a shared gate. We study separate networks with local routers and no cross-network gradients, and measure specialization as the alignment of expert identity with latent structure rather than inferring it from routing statistics.

\paragraph{Populations of separately trained models.}
A separate body of work trains full models independently and composes them afterwards: branching and merging domain experts~\citep{li2022branch,sukhbaatar2024branch}, discovering domains by clustering before training~\citep{gururangan2023scaling}, heterogeneous domain ensembles~\citep{ersoy2025hdee}, weight averaging and task arithmetic~\citep{wortsman2022modelsoups}, routing among adapter libraries~\citep{ostapenko2024towards,muqeeth2024learning}, and decentralized path composition~\citep{douillard2024dipaco}. In these systems the data partition is fixed in advance or expertise is composed only after training; our agents learn during training, from an unlabeled mixture, who handles what, and exchange information at inference time.

\paragraph{Communication between separate networks.}
Learned communication has been studied in multi-agent reinforcement learning~\citep{sukhbaatar2016learning,foerster2016learning,jiang2018learning,das2019tarmac}, where roles can emerge among homogeneous agents~\citep{wang2020roma,mordatch2018emergence,baker2019emergent}, and in collaborative perception, where agents share intermediate features and learn when and with whom to communicate~(\citealp{liu2020when2com,hu2022where2comm}; \citealp{wang2023core}). Closest in mechanism to ours are systems in which one network reads another's internal states through cross-attention, as in CALM~\citep{bansal2024llm}, and key--value communication between language models~(\citealp{fu2026cache}; \citealp{ye2025kvcomm}). These fix sender and receiver in advance and train a bridge for the pair. We ask whether a population can discover who should communicate with whom from the improvement a message brings, and whether keys and values are useful when the sender receives no gradient from the receiver; our agents also see the same input, so gains cannot come from fusing complementary views. No prior work isolates the combination studied here: identical agents, one unlabeled mixture, one shared self-supervised objective, local routing, and stop-gradient latent communication, with metrics that separate specialization from utilization.

\section{A Controlled Testbed for Emergent Specialization}\label{sec:method}

\subsection{Measuring Specialization and Utilization}\label{sec:metrics}

We assume that data are generated by a latent factor model: there is a latent variable \( y \in \{1, \dots, M\} \) with prior \( p(y) \) such that \( p_{\rm D}(x) = \sum_{y} p(x \mid y)\, p(y) \). Consider a population of \( K \) agents \( \mathcal{P}_K = \{a_i\}_{i=1}^K \), each evaluated on an input \( x \) through a loss \( l(x, a_i) \). We call the agent with the lowest loss the \emph{expert} on \( x \), \( e(x) = \arg\min_{1 \le i \le K} l(x, a_i) \), which assigns responsibility for each input to one agent. The population is \emph{specialized} if this assignment correlates with \( y \). We quantify the dependence with normalized mutual information, which is comparable across numbers of agents and factors,
\begin{align}
    \mathrm{Sp}(\mathcal{P}_K)
    = \mathrm{NMI}(e(x), y)
    = \frac{\mathrm{MI}(e(x), y)}{\min\{H(e(x)), H(y)\}} \in [0,1].
\end{align}
If one agent is the expert everywhere, \( \mathrm{Sp}=0 \). However, \( \mathrm{Sp} \) can be close to 1 when one agent handles almost all inputs and the others own tiny niches. The \emph{utilization} \( \mathrm{U}(\mathcal{P}_K) = H(e(x)) / \ln K \in [0,1] \), the normalized entropy of the expert assignment, separates a balanced division of labor from such collapse. A well-specialized population has high \( \mathrm{Sp} \) \emph{and} high \( \mathrm{U} \). Both are estimated on a validation set where \( y \) is known and depend only on the agents' standalone losses, not on any router, so a balanced router does not by itself produce high \( \mathrm{Sp} \): the distinction that routing statistics alone cannot make~\citep{wang2026illusion}.

\subsection{Experimental Setting, and Why We Chose It}\label{sec:setting}

\paragraph{Objective.} Given an RGB image, a frozen DINOv3 encoder~\citep{simeoni2025dinov3} produces patch features \( z \in \mathbb{R}^{d \times h \times w} \). We sample a mask \( m \in \{0,1\}^{h \times w} \) that keeps a random rectangular window of 30\% of the patches visible and hides the remaining 70\%, and ask an agent to reconstruct the full feature map from \( \hat z = m \cdot z \). Writing \( x = (\hat z, z) \) for the resulting training input, the loss is the cosine distance on the masked positions, \( l(x, a_i) = D_{\rm cos}(z, a_i(\hat z)) \). All agents are small ViTs~\citep{dosovitskiy2020image} with 4--8 layers (Appendix~\ref{app:architectures}).

\paragraph{Performance measure.} Cosine distance is a natural training loss but a poor performance measure, because outpainting is ambiguous. Instead of downstream probing~\citep{dinowm}, we use a lightweight retrieval proxy: for each predicted feature at a hidden position we retrieve its five nearest neighbors (cosine) among the 196 ground-truth patch features of the same image and count a hit if the true position is among them; \emph{R-Top5} is the hit rate over hidden positions, averaged over images (trivial baselines in Appendix~\ref{app:eval_protocol}). For a population we report the expert envelope over the validation set, \( \mathrm{R\mbox{-}Top5\mbox{-}Best} = \frac{1}{|D_{\rm val}|} \sum_{x \in D_{\rm val}} \mathrm{R\mbox{-}Top5}(x, a_{e(x)}) \), the score of the expert (lowest-loss agent) on each input, and \( \mathrm{R\mbox{-}Top5\mbox{-}Worst} \), the score of the highest-loss agent. Best is the value the population would attain if every input reached its expert, a routing upper bound at fixed agents. It is selected by loss and scored by R-Top5, so it is not the metric's maximum over agents; Appendix~\ref{app:eval_protocol} reports how often the two coincide. Worst measures how much a non-expert loses. The performance actually attained lies in between and is reported at two operating points (Table~\ref{tab:main}): the \emph{delegation} performance evaluates each sample with the agent that the regime's routing rule selects, and the \emph{collaboration} performance lets a randomly chosen agent reconstruct it with the help of that selected agent, reading its keys and values. In the decentralized regimes the rule is the drawn agent's own router (Section~\ref{sec:regimes}).

\paragraph{Data.} A single model is pretrained on ImageNet100~\citep{deng2009imagenet}. All agents are then initialized from it and fine-tuned with LoRA~\citep{hu2022lora} on a mixture of six datasets that are out of distribution with respect to pretraining: AID~\citep{xia2017aid} (aerial), KITTI~\citep{geiger2013vision} (driving), BridgeDataV2~\citep{walke2023bridgedata} and RT-1~\citep{rt12022arxiv} (robot manipulation), CelebA~\citep{liu2015faceattributes} (faces), and 102 Flowers~\citep{Nilsback08}. By default we use 5,999 training and 2,190 validation images per dataset (35,994 and 13,140 in total; split construction and sequence handling in Appendix~\ref{app:data}). An imbalanced mixture \( D_{\rm imb} \) with 4K, 4K, 8K, 9K, 7K, and 4K training images keeps the total of 36K fixed. The dataset index is the latent factor \( y \) used for evaluation and is never shown to any agent or router.

\paragraph{Why this setting.} Each choice determines what the results can say.
\emph{One shared self-supervised objective.} Different losses or rewards would create roles by design; one loss on one unlabeled mixture is what makes a division of labor \emph{emergent}.
\emph{Prediction in a frozen semantic space.} Like next-frame prediction and denoising, masked feature prediction infers missing content from context under a single loss. Working in DINOv3 space removes pixel-level ambiguity, lets a 4--8 layer ViT capture meaningful structure, and makes a population trainable in about six hours on four GPUs (Appendix~\ref{app:training}), so that regimes, population sizes, capacities, and mixtures can be swept.
\emph{A known latent factor.} Measuring specialization needs ground truth: six visually distinct domains outside the pretraining distribution give a factor \( y \) that is well defined, unseen during pretraining, and hidden during fine-tuning.
\emph{Whole-sample routing among separate networks} keeps every agent a standalone learner and makes specialization interpretable at the level of the data.

\begin{table}[t]
\centering
\caption{The ladder of training regimes. Each step removes one form of central control. All population regimes share the same $K$ agents, backbones, initialization, LoRA budget, and total training samples.}
\label{tab:roadmap}
\footnotesize
\setlength{\tabcolsep}{4pt}
\begin{tabularx}{\linewidth}{@{}l l >{\raggedright\arraybackslash}X c >{\raggedright\arraybackslash}X@{}}
\toprule
Regime & Router & Router signal & Training & Role in the study \\
\midrule
Solo & -- & -- & -- & Generalist reference \\
Random & central, fixed & none (random split) & non-local & Control: population without semantics \\
K-means & central, fixed & clusters of clean DINOv2 features & non-local & Positive control: privileged split \\
Learned central & central, trained & policy gradient on helper loss & non-local & MoE-style reference: one central gate trained online \\
Distr.\ delegation & one per agent & policy gradient on helper loss & non-local & Q2: no central router \\
\methodName (ours) & one per agent & policy gradient on \emph{improvement} & local & Q3: is expertise shareable? \\
\bottomrule
\end{tabularx}
\end{table}

\subsection{A Ladder of Training Regimes: From Central Control to Local Collaboration}\label{sec:regimes}

The regimes in Table~\ref{tab:roadmap} lead from full central control to the target setting. The first four are diagnostic controls built from standard components. Distributed delegation removes the central router, but a sample still has to be passed to the agent that solves it best. \methodName additionally lets the requester reconstruct the sample itself \emph{with} the selected agent's help. In every population regime a \emph{router} \( r \colon x \mapsto \{1, \dots, K\} \) maps its input to the index of an agent, and each agent is trained on the samples routed to it. The input to the router is the corrupted sample \( \hat z \) in every learned regime, and the clean features only for the K-means control. The reconstruction loss of a batch \( \{x_i\}_{i=1}^b \) with routing \( r \) is \( \mathcal{L}_{\rm rec} = \frac{1}{b}\sum_{i = 1}^b l(x_i, a_{r(x_i)}) \), so that routing changes the effective data distribution seen by each agent and is the only mechanism through which specialization can arise.

\paragraph{Solo, random, and K-means routing.} The \emph{solo} model is one agent trained on the full mixture, with no routing. \emph{Random} routing splits the training set into \( K \) random fixed subsets and trains each agent on its own subset: the population has \( K \) agents and the same data budget but no semantic assignment. It isolates the effect of having several models from the effect of specialization. \emph{K-means} routing clusters the training set offline into \( K \) groups, using the \texttt{[CLS]} features that a frozen DINOv2 ViT-B/14 encoder~\citep{oquab2024dinov2} computes on the \emph{uncorrupted} images, and assigns each group to an agent. It is a positive control with privileged information.

\paragraph{Learned central routing.} A trainable router \( r_\theta \) (a two-layer MLP on the mean of the \emph{visible} DINO tokens; it sees only what an agent sees) outputs a distribution over agents. For sample \( i \) the assigned agent \( j \) is sampled, \( j \sim r_\theta(\hat z_i) \). Since sampling is not differentiable, the router is trained with a policy-gradient loss added to the reconstruction loss,
\begin{align}\label{eq:pg}
    \mathcal{L}_{\rm router}
    = - \frac{1}{b} \sum_{i = 1}^b \left(R_i - \bar R\right) \log r^{(j)}_\theta(\hat z_i),
    \qquad
    R_i = -{\rm sg}\left[l(x_i, a_j)\right],
\end{align}
where \( \bar R \) is an exponential moving average of the rewards and \( {\rm sg}[\cdot] \) denotes the stop-gradient operator. This is the closest analogue of an MoE gate in our setting: one online router, trained jointly with the agents, but routing whole samples between separate networks rather than tokens between sub-modules~\citep{jacobs1991adaptive,fedus2022switch}. All learned routers in this paper additionally receive a small KL penalty toward a uniform marginal routing distribution (Appendix~\ref{app:method}), the standard load-balancing mechanism of MoEs; Section~\ref{sec:res_robust} reports what happens without it.

\paragraph{Distributed delegation: removing the central router.} We now give every agent its own routing head. For each training batch a \emph{requester} \( a_q \) is drawn uniformly at random. Its router assigns each sample \( i \) in the batch to a \emph{helper} \( j \sim r_q(\hat z_i) \), which is trained on the sample as before, and \( r_q \) is trained with Eq.~\eqref{eq:pg}. Each head is a two-layer MLP on the mean of the requester's own queries, so a router reuses its agent's features but has its own parameters. The requester is excluded from its own routing distribution (Section~\ref{sec:res_robust} lifts this exclusion). No central component is left, but the learning signal is still not local: a router is rewarded by \emph{another} agent's loss, and the requester only finds an agent that solves the sample better, so the population can never do better than its agents do alone on their niches.

\begin{figure}[!t]
    \centering
    \includegraphics[width=0.8\linewidth, trim=1.2cm 1.8cm 1.6cm 0.3cm, clip]{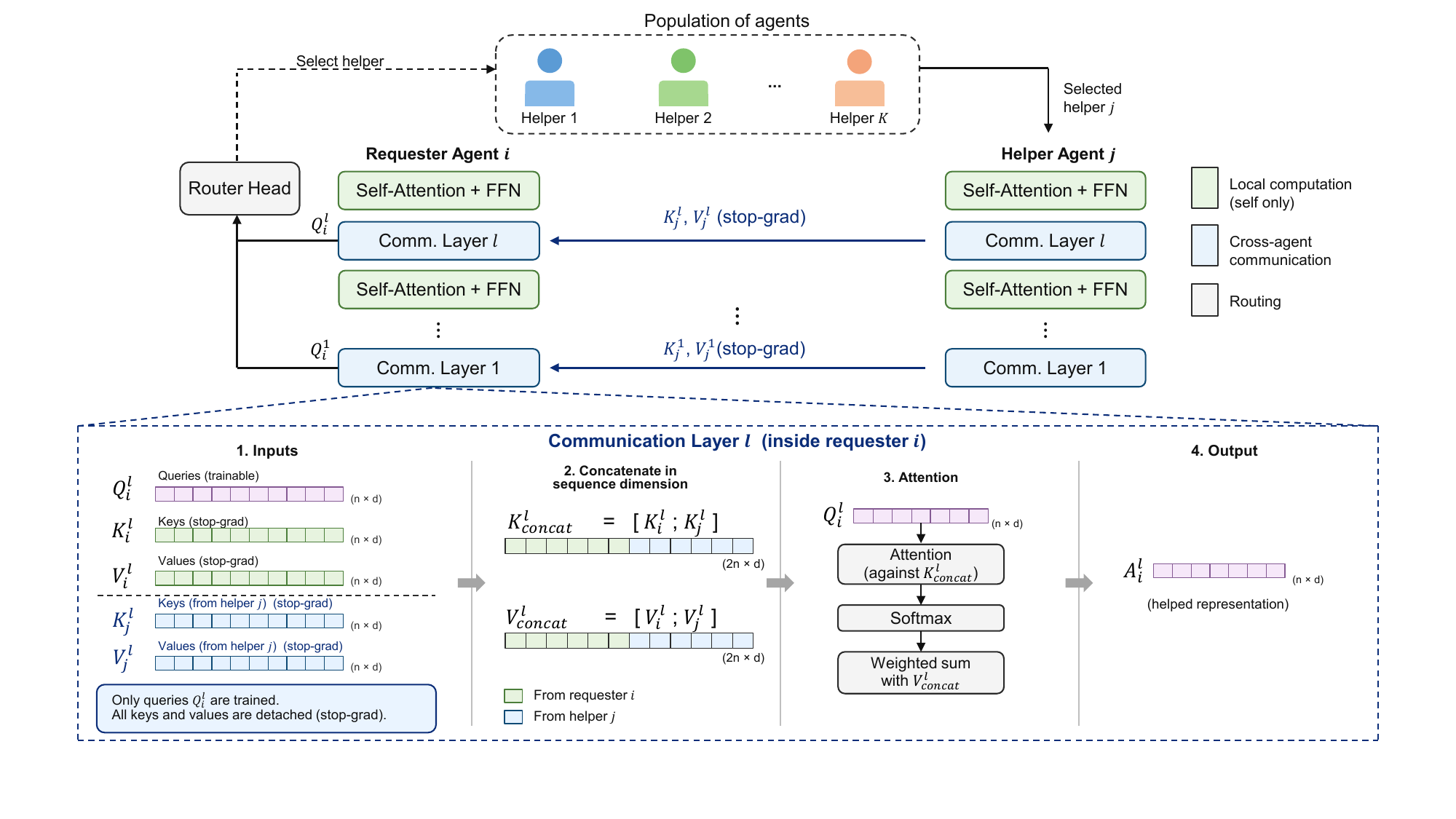}
    \caption{\methodName. A random requester routes each sample to a helper with its own router; in communication layers its queries attend to its own and the helper's detached keys and values. The helper is trained on the sample, the requester on its helped reconstruction, the router on the improvement.}
    \label{fig:disco}
\end{figure}

\paragraph{\methodName: distributed collaboration.} We want the requester to \emph{use} the helper rather than hand the sample over, so that expertise can flow across the population. \methodName therefore changes two things with respect to delegation, the channel and the router reward, and Section~\ref{sec:res_robust} separates their effects. Training the requester on every sample as well would restore the symmetry between agents and remove the specialization. Instead, the helper communicates part of its internal computation and the requester learns to reconstruct \emph{with} that message. Two design decisions define the protocol.

\emph{(1) Messages are keys and values with stopped gradients.} A differentiable channel that backpropagates the requester's loss into the helper would train helpers to produce useful messages, but it would also couple all agents into one optimization graph, the MoE regime we depart from. We therefore stop all gradients through the channel and build messages from quantities the helper already computes for itself. The \( l \)-th self-attention block of agent \( a_i \) outputs \( A_{\rm solo}(a_i, l) = {\rm SoftMax}(Q_i^l K_i^{l\top}/\sqrt{d})\, V_i^l \) with \( Q_i^l, K_i^l, V_i^l \in \mathbb{R}^{n \times d} \). In a \emph{communication layer}, after requester \( a_i \) has selected helper \( a_j \), the helper's keys and values are concatenated to the requester's own:
\begin{align}
    A_{\rm helped}(a_i, a_j, l)
    = {\rm SoftMax}\!\left(\frac{Q_i^l K_{\rm cat}^{l\,\top}}{\sqrt{d}} \right) V_{\rm cat}^l,
    \quad
    K_{\rm cat}^{l} = {\rm sg}\!\left[K_i^{l};\, K_j^{l}\right],\;
    V_{\rm cat}^{l} = {\rm sg}\!\left[V_i^{l};\, V_j^{l}\right],
\end{align}
where \( [\cdot;\cdot] \) concatenates along the token axis. The requester is free to ignore the helper's entries, and communication can be switched on per layer (Table~\ref{tab:architecture}). A requester thus has a solo loss \( l_{\rm solo}(x, a_i) \) and a helped loss \( l_{\rm helped}(x, a_i, a_j) \). Figure~\ref{fig:disco} illustrates the pipeline.

\emph{(2) Routers are rewarded by improvement, not by the helper's accuracy.} In delegation the router learns who is best in isolation. In \methodName the router of requester \( a_q \) is trained with Eq.~\eqref{eq:pg} using \( R_i = {\rm sg}[\, l_{\rm solo}(x_i, a_q) - l_{\rm helped}(x_i, a_q, a_j) \,] \), the reduction in the requester's own loss due to the help. This signal is local to the requester and never uses the helper's standalone performance. Section~\ref{sec:res_routing} asks whether such routers nonetheless find the experts. The full objective is
\begin{align}
    \mathcal{L}_{\rm \methodName}
    = \lambda_{\rm rec} \mathcal{L}_{\rm rec}
    + \lambda_{\rm router} \mathcal{L}_{\rm router}
    + \lambda_{\rm helped} \mathcal{L}_{\rm helped},
    \qquad
    \mathcal{L}_{\rm helped} = \frac{1}{b} \sum_{i = 1}^b l_{\rm helped}(x_i, a_q, a_j),
\end{align}
where \( \mathcal{L}_{\rm rec} \) trains the helpers on their assigned samples as in all other regimes. Algorithm~\ref{alg:disco} in Appendix~\ref{app:method} summarizes one step. Every component is standard (ViTs, LoRA, attention, REINFORCE); what is specific to \methodName is their combination, local improvement-based routing with stop-gradient key--value communication.

\section{Results}\label{sec:experiments}

In this section, we follow the ladder of Table~\ref{tab:roadmap}, one question per subsection. Unless stated otherwise, populations have \( K=4 \) type-B agents, i.e., base size with 43M parameters each (Table~\ref{tab:architecture}), are trained on the balanced mixture, and all regimes see the same number of training samples. All numbers are means over three draws of crops and masks on the validation set, and the intervals we quote are paired bootstrap intervals over images (Appendix~\ref{app:eval_protocol}). They measure evaluation-set uncertainty for the trained checkpoints, not variability across training runs, which Section~\ref{sec:res_robust} addresses separately.

\begin{table}[t]
\centering
\caption{Routing study. Best/Worst is the expert envelope (R-Top5 of the lowest-/highest-loss agent). Deleg./Collab.: performance with the regime's own routing rule, without and with communication, for a random requester. Worst+help: the worst agent on an input reconstructing with the selected agent. Means over three draws of crops and masks shared by all regimes; 95\% paired bootstrap intervals within \( \pm0.003 \) per entry and \( \pm0.0015 \) for differences between regimes (Appendix~\ref{app:eval_protocol}).}
\label{tab:main}
\footnotesize
\setlength{\tabcolsep}{6pt}
\begin{tabular}{@{}l cc cc ccc@{}}
\toprule
Regime & Sp\(\uparrow\) & U\(\uparrow\) & Best\(\uparrow\) & Worst\(\uparrow\) & Deleg. & Collab. & Worst+help \\
\midrule
Solo & -- & -- & 0.653 & 0.653 & -- & -- & -- \\
Random & 0.011 & 0.999 & 0.650 & 0.634 & 0.642 & 0.643 & 0.637 \\
\midrule
K-means & \best{0.957} & \second{0.962} & \best{0.678} & 0.535 & \best{0.678} & 0.618 & 0.582 \\
Learned central & 0.788 & \best{0.995} & \best{0.678} & 0.533 & \best{0.678} & 0.620 & 0.592 \\
Distr.\ delegation & 0.926 & 0.846 & \second{0.672} & \second{0.546} & \second{0.665} & \second{0.626} & \second{0.593} \\
\methodName & \second{0.931} & 0.841 & 0.668 & \best{0.568} & 0.662 & \best{0.653} & \best{0.644} \\
\bottomrule
\end{tabular}
\end{table}

\begin{figure}[t]
\centering
\begin{minipage}[t]{0.33\linewidth}
\centering
\vspace{0pt}\includegraphics[width=\linewidth]{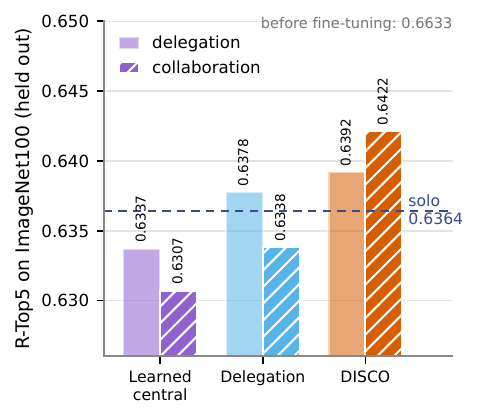}
\captionof{figure}{Retention on the ImageNet100 validation set (5,000 images, seen during pretraining, unseen during specialization; protocol of Table~\ref{tab:main}). All populations lose accuracy with respect to the model before fine-tuning. \methodName with communication loses least and is the only regime in which communication helps.}
\label{fig:helper_helped}
\end{minipage}\hfill
\begin{minipage}[t]{0.64\linewidth}
\centering
\vspace{0pt}\captionof{table}{Same requester, same input, different message. R-Top5 of a requester reconstructing with the message in the row, averaged over the four requesters (``all''), or over the requesters that are not the expert on the input (``non-exp.''). Selected helper is the Collab.\ column of Table~\ref{tab:main}. Three draws; 95\% paired intervals on differences between rows are within \( \pm0.002 \).}
\label{tab:interventions}
\footnotesize
\setlength{\tabcolsep}{3pt}
\begin{adjustbox}{max width=\linewidth}
\begin{tabular}{@{}l cc cc@{}}
\toprule
 & \multicolumn{2}{c}{\methodName} & \multicolumn{2}{c}{Delegation} \\
\cmidrule(lr){2-3} \cmidrule(lr){4-5}
Message & all & non-exp. & all & non-exp. \\
\midrule
alone (no message) & 0.618 & 0.601 & 0.602 & 0.578 \\
selected helper & 0.653 & 0.649 & 0.626 & 0.616 \\
expert (lowest loss) & 0.654 & 0.649 & 0.630 & 0.616 \\
random other agent & 0.636 & 0.628 & 0.609 & 0.597 \\
own K/V duplicated & 0.618 & 0.601 & 0.602 & 0.578 \\
selected, other image's K/V & 0.380 & 0.366 & 0.404 & 0.403 \\
all-zero K/V & 0.608 & 0.590 & 0.594 & 0.570 \\
\bottomrule
\end{tabular}

\end{adjustbox}
\end{minipage}
\end{figure}

\subsection{Q1: Does Specialization Matter, and Is It Specialization or Just More Models?}\label{sec:res_central}

As Table~\ref{tab:main} shows (plotted in Appendix~\ref{app:dynamics}), central routing, privileged or learned, produces specialized, well-utilized populations (\( \mathrm{Sp} = 0.79 \)--\( 0.96 \), \( \mathrm{U} = 0.96 \)--\( 1.00 \)) and raises R-Top5-Best from \( 0.653 \) (solo) to \( 0.678 \). The K-means control specializes most, while the learned router, which sees only the masked input, recovers the full gain online with a less semantic split.

The random-routing control is what gives these numbers their meaning. It uses the same four agents, backbones, and data budget, and its best-agent responsibility is balanced (\( \mathrm{U}=1.00 \)), but it carries no semantic structure (\( \mathrm{Sp}=0.01 \)) and performs \emph{worse} than the solo model, since each agent sees a quarter of the data: several models are not enough, and balanced load is not specialization. The gain appears only when responsibility aligns with latent structure. At the same time the Worst endpoint drops from \( 0.653 \) to \( \approx 0.53 \): a specialized agent is a poor generalist, which is why sharing expertise (Q3) matters.

\subsection{Q2: Does Specialization Survive the Removal of the Central Router?}\label{sec:res_delegation}

Distributed delegation replaces the single gate by one router per agent, each trained only on the batches for which its agent is the requester. Specialization nevertheless emerges (\( \mathrm{Sp}=0.93 \)), close to the K-means split and above the learned central router (\( 0.79 \)), and R-Top5-Best (\( 0.672 \)) is within 0.006 of the central regimes. Utilization is lower (\( \mathrm{U}=0.85 \)): without a global view the population balances responsibility less evenly. Nothing in the local routers' objective refers to domains, yet the population is almost as domain-aligned as an offline clustering of clean features, grouping the six domains into three two-domain niches with one agent almost idle (Appendix~\ref{app:assignment}).

\subsection{Q3: Can Emergent Expertise Be Shared Through a Stop-Gradient Channel?}\label{sec:res_disco}

\methodName specializes as much as delegation (\( \mathrm{Sp}=0.93 \), \( \mathrm{U}=0.84 \); niches in Appendix~\ref{app:assignment}) at a slightly lower expert Best (\( 0.668 \) vs.\ \( 0.672 \)). First, \methodName has the highest R-Top5-Worst of the specialized populations (\( 0.568 \) vs.\ \( 0.533 \)--\( 0.546 \)). Second, communication recovers a non-expert to generalist level: a random requester reconstructing with its selected helper reaches \( 0.653 \), on par with the solo model and \( 0.027 \) above the same measurement in delegation (\( 0.626 \)). The same requester alone scores \( 0.618 \), so the message is worth \( 0.035 \) to it. Collaboration does not beat delegating to the expert on the mixture (\( 0.653 \) vs.\ \( 0.662 \)); its value is that a non-expert becomes as good as a generalist while the experts remain better than one. The worst agent on an input, once helped, improves from \( 0.568 \) to \( 0.644 \) (delegation: \( 0.546 \) to \( 0.593 \); Figure~\ref{fig:reconstructions} shows what the help restores). Third, on the ImageNet100 validation set, which the agents saw during pretraining but not during specialization (a retention test rather than transfer to a new distribution), every population and the solo model fall below the model before fine-tuning (\( 0.663 \)). \methodName with communication loses least (\( 0.642 \) vs.\ \( 0.636 \) solo, \( 0.634 \) delegation, \( 0.631 \) learned central; paired difference to solo \( +0.006 \), interval \( [0.005, 0.006] \)), and it is the only regime in which communication raises performance above the regime's own delegation (Figure~\ref{fig:helper_helped}). All differences in this section are resolved by the paired intervals (Appendix~\ref{app:eval_protocol}).

\subsection{Q4: What Governs When Specialization Emerges?}\label{sec:res_scaling}

\paragraph{Population size.} We train \methodName with 4, 6, and 8 agents (Figure~\ref{fig:scaling}a,b, Appendix~\ref{app:dynamics}). All populations reach high specialization and larger ones end with higher R-Top5-Best but specialize \emph{later}: requesters are drawn uniformly, so each router is trained on fewer batches and the symmetry takes longer to break.

\paragraph{Agent capacity.} With S, B, and L agents (14M, 43M, 103M parameters; Table~\ref{tab:architecture}) in populations of four, performance grows with capacity, and more capable agents specialize \emph{earlier} (Figure~\ref{fig:scaling}c,d). Larger agents produce more informative queries, keys, and values, so the improvement signal tells helpers apart sooner.

\paragraph{Data imbalance.} On the imbalanced mixture \( D_{\rm imb} \), specialization is somewhat weaker (\( \mathrm{Sp}=0.75 \)), utilization higher (\( \mathrm{U}=0.93 \)), performance slightly lower, and the grouping of domains changes (Appendix~\ref{app:assignment}); the population does not collapse onto the dominant domains.

\subsection{Q5: Do Local Routers Find the Experts?}\label{sec:res_routing}

The population \emph{agrees} on an input if the most probable helper \( h_i(\hat z) = \arg\max_{j} r_i^{(j)}(\hat z) \) of every non-expert requester \( i \neq e(x) \) is the expert. The 4-agent \methodName population agrees on \( 98.3\% \) of the validation inputs (delegation: \( 97.9\% \); 6 agents \( 98\% \), 8 agents \( 88\% \), Table~\ref{app:scaling_table}). The routers never saw a helper's standalone loss; that they converge on the loss-defined expert means that ``most helpful to me'' and ``best at this input'' coincide, so the division of labor is visible to the agents themselves (attention patterns in Appendix~\ref{app:assignment}).

\subsection{Q6: Ablations and Robustness}\label{sec:res_robust}

\paragraph{Same requester, different messages.} Table~\ref{tab:interventions} evaluates every requester on every validation input while only the message changes. A \methodName requester scores \( 0.618 \) alone, \( 0.653 \) with the helper its router selects, and \( 0.636 \) with a random other agent's message. Information-free messages of the same shape isolate the content: duplicating the requester's own keys and values changes nothing, all-zero keys and values cost \( 0.010 \), and the selected helper's keys and values computed for a \emph{different} image cost \( 0.239 \): the requester reads the message. Non-expert requesters gain \( 0.048 \) while the expert loses \( 0.004 \). Delegation agents, never trained to read a message, show the same ordering with smaller values.

\paragraph{Training seeds at a quarter of the budget.} Appendix~\ref{app:seeds} repeats the two distributed regimes with three new seeds for 100 fine-tuning epochs. Specialization emerges in every run (\( \mathrm{Sp}=0.92 \)--\( 0.95 \) for \methodName, \( 0.86 \)--\( 0.95 \) for delegation), the ladder of Section~\ref{sec:res_central} replicates, and in every seed \methodName trades an expert Best within \( 0.002 \) of delegation for a higher Worst (\( 0.594 \) vs.\ \( 0.568 \)), collaboration (\( 0.631 \) vs.\ \( 0.622 \)), and Worst+help (\( 0.625 \) vs.\ \( 0.607 \)). What varies is the shape of the division of labor: utilization ranges from 0.85 to 0.96 in both regimes, and whether \methodName specializes \emph{more} than delegation is seed-dependent (\( 0.94\pm0.02 \) vs.\ \( 0.89\pm0.05 \)) while its advantage in Worst and in the helped scores is not.

\paragraph{Reward versus channel.} Two runs at the same budget separate the two changes \methodName makes to delegation (Table~\ref{tab:seeds}). Training the requester on its helped reconstruction while keeping delegation's router reward reproduces the helped scores of \methodName (collaboration \( 0.631 \), Worst+help \( 0.624 \), Worst \( 0.586 \)) with delegation's specialization (\( \mathrm{Sp}=0.86 \)) and routing agreement (82\% vs.\ 98\%): learning to read the message, not the improvement reward, creates the gain of a helped requester, while the improvement reward is what aligns the routers with the expert. Communication training under random routing yields neither specialization (\( \mathrm{Sp}=0.00 \)) nor a benefit from messages (\( 0.627 \) helped vs.\ \( 0.626 \) alone): messages are worth reading only once routing has made the agents different.
\paragraph{Protocol choices.} Without the KL term toward a uniform routing marginal (\( \lambda_{\rm KL}=0 \)), the local routers of \methodName collapse within four epochs onto one agent, which becomes a generalist expert on 90\% of the inputs (\( \mathrm{Sp}=0.08 \), \( \mathrm{U}=0.25 \), Best \( 0.633 \), the solo level): the load-balancing pressure that MoEs also use is a precondition for the division of labor, while the random-routing control shows that balance alone does not produce one; what emerges is the alignment of the balanced niches with the domains. Allowing a requester to select itself slows and weakens specialization (\( \mathrm{Sp}=0.81 \) after 72 epochs, all four agents used) but does not prevent it: a router selects its own agent for 91\% of the inputs on which that agent is the expert and for only 2\% of the others, and the helped scores equal those of \methodName.

\section{Discussion}\label{sec:discussion}

We studied emergent specialization in populations of initially identical neural agents trained with a shared self-supervised reconstruction objective, without task labels, predefined roles, or a central orchestrator. Using specialization and utilization as metrics, we showed that agents organize around the latent structure of heterogeneous visual data as soon as routing is tied to per-sample performance, and that this improves on both a single agent and a non-specialized population of the same size and budget. What a router is trained on, rather than how evenly it spreads the load, makes the division of labor semantic, and simple local routing with constrained communication suffices to support it.

We introduced \methodName, a distributed collaborative routing framework in which agents locally select helpers and communicate through attention keys and values with stopped gradients. Across our experiments \methodName induces strong specialization, is the only regime of Table~\ref{tab:main} in which a requester gains from the message it receives, and remains effective across population size, agent capacity, and data balance. Future work involves larger, more heterogeneous populations, richer messages, and other predictive objectives.


\section*{Acknowledgements}

Aram Davtyan, Pablo Acuaviva and Sebastian Stapf have been supported by
Swiss National Science Foundation (SNSF) Project 10001278.

\bibliographystyle{plainnat}
\bibliography{common/references}

\newpage
\appendix
\section{Implementation details}\label{app:implementation_details}

\subsection{Agent architectures}\label{app:architectures}

All agents are small ViT networks~\citep{dosovitskiy2020image}. Table~\ref{tab:architecture} lists the three sizes. Communication layers are the layers in which a requester's queries may attend to a helper's keys and values. In all other layers every agent attends only to itself.

\begin{table}[!htbp]
    \centering
    \caption{Architecture details of agents of different sizes.}
    \label{tab:architecture}
    \begin{tabularx}{\linewidth}{@{}r|*{3}{>{\centering\arraybackslash}X}@{}}
    \toprule
         & S & B & L \\
    \midrule
        Model dim & 512 & 768 & 1024 \\
        Number of attention heads & 8 & 12 & 16 \\
        Number of layers & 4 & 6 & 8 \\
        Feed-forward hidden dimension & 2048 & 3072 & 4096 \\
        Dropout & 0.05 & 0.05 & 0.05 \\
        Communication layers & [1, 3] & [1, 2, 4, 5] & [1, 2, 3, 5, 6, 7] \\
        Number of parameters & 13.7M & 43.1M & 103M \\
    \bottomrule
    \end{tabularx}
\end{table}

\subsection{Routers}

Table~\ref{tab:routers} makes explicit what each router receives and how it is trained, since this determines what each regime can and cannot exploit. All learned routers are two-layer MLP heads with their own parameters: LayerNorm, a linear layer from the agent's model dimension \( d \) to \( d \), GELU, and a linear layer to \( K \) logits, i.e., 0.27M, 0.60M, and 1.06M parameters for S, B, and L agents. The logits are divided by the temperature \( \tau=1.25 \) before the softmax. In the central regime the head is applied to the mean of the \emph{visible} DINO tokens of the masked input, so it has exactly the information available to an agent. In the distributed regimes each agent \( a_q \) owns a head \( r_q \) applied to the mean of that agent's queries in its communication layers. The router therefore reuses the agent's (LoRA-adapted) feature computation, but the MLP parameters are not shared between agents. Distributed delegation and \methodName read the queries of the \emph{same} layers, so that the two regimes differ only in the router's training signal and not in what the router sees. The K-means router clusters the \texttt{[CLS]} features of the \emph{uncorrupted} training images offline (computed with a frozen DINOv2 ViT-B/14, i.e., an encoder different from the DINOv3 target space). It is the only router with access to unmasked inputs and is used as a positive control.

\begin{table}[!htbp]
\centering
\caption{Information and training signal of each router.}
\label{tab:routers}
\small
\begin{tabularx}{\linewidth}{@{}l X X X@{}}
\toprule
Regime & Router input & Parameters & Training signal \\
\midrule
Random & none & none (fixed random split of the training set) & none \\
K-means & DINOv2 \texttt{[CLS]} feature of the \emph{clean} image & none (offline clustering, $K$ clusters) & none \\
Learned central & mean of visible DINO tokens of the masked input & one shared 2-layer MLP & REINFORCE, reward $=-$ helper loss \\
Distr.\ delegation & mean of requester $a_q$'s queries in communication layers & one 2-layer MLP per agent & REINFORCE, reward $=-$ helper loss; requester excluded \\
\methodName & same as delegation & one 2-layer MLP per agent & REINFORCE, reward $=$ requester's improvement $l_{\rm solo}-l_{\rm helped}$; requester excluded \\
\bottomrule
\end{tabularx}
\end{table}

\subsection{Training details}\label{app:training}

All models are pretrained on ImageNet100 for 100 epochs with 64 images per GPU on four GPUs (global batch size 256) using AdamW~\citep{loshchilov2017decoupled} with learning rate $10^{-4}$ and weight decay $10^{-4}$. The pretrained model is a single agent of the target size trained with the same objective as the populations (a random 30\% visible window, cosine loss on the hidden positions) and with all of its weights trainable. AdamW uses \( \beta=(0.9, 0.999) \) and \( \epsilon=10^{-8} \); the learning rate is constant, without warm-up or decay, and gradients are clipped to norm 1.0. Images are augmented with a random resized crop (scale 0.6--1.0) to \( 224\times224 \) and a horizontal flip; DINOv3 ViT-B/16 then yields a \( 14\times14 \) grid of 196 patch tokens of dimension 768 (the class and register tokens are discarded), which are L2-normalized before an agent sees them. Each agent maps these tokens to its model dimension with a linear layer (the identity for type B, whose model dimension is 768), replaces the hidden positions by a learned mask token, adds a fixed sinusoidal positional encoding over the 196 positions, applies pre-LayerNorm transformer blocks with GELU feed-forward layers and a final LayerNorm, and maps back to 768 dimensions with a linear layer. All populations are trained on 4 NVIDIA GeForce RTX 4090 GPUs. The four data-parallel processes each draw their own requester, so one optimizer step involves four requesters with 64 samples each. Fine-tuning uses the same batch size and optimizer. The populations are fine-tuned on the mixture for 400 epochs using LoRA~\citep{hu2022lora} with rank 64, $\alpha = 8$ and no dropout. LoRA adapters are inserted in the query, key, value, output, and feed-forward projections of the communication layers only (Table~\ref{tab:architecture}). All other agent weights, including the non-communication layers, stay frozen, and the routing head is trained in full. The adapters account for 1.18M, 3.54M, and 7.08M parameters per S, B, and L agent and the routing head for 0.27M, 0.60M, and 1.06M, i.e., 1.45M, 4.13M, and 8.13M trainable parameters per agent and 16.5M for a four-agent type-B population. The solo model uses the same adapters (3.54M); its routing head has no decision to make and receives no learning signal. Random and K-means routing train no router, learned central routing trains one head, and the distributed regimes one head per agent. Fine-tuning four type-B agents takes 6.1 hours with \methodName, 5.0 hours with distributed delegation, and 5.3 hours with learned central routing. The solo model takes 4.6 hours. Part of the \methodName overhead is due to additional forward passes used only for logging (e.g., helped reconstructions); Appendix~\ref{app:compute} counts the passes each regime actually needs. All regimes are trained with the same random seed and data order, so that differences between regimes cannot be attributed to initialization or sample order; Section~\ref{sec:res_robust} varies the seed.

\subsection{Data construction}\label{app:data}

The mixture takes the first 5,999 training and the first 2,190 validation images of every source in the order in which the loader enumerates it (class folders for AID, episode order for the robot datasets, directory order otherwise); 5,999 and 2,190 are the sizes of the smallest source, 102 Flowers. The two robot datasets are stored as episodes. Training and validation use disjoint sets of episodes, an episode is never split across the two, and only every fifth frame of an episode is used, so the 5,999 training frames come from 659 BridgeDataV2 and 559 RT-1 episodes and the 2,190 validation frames from 245 and 203 episodes. KITTI uses the left colour camera of the object benchmark: its training set for training and its test set for validation, which come from different drives. No near-duplicate filtering beyond this frame subsampling was applied, so strided frames of one episode remain similar within a split; this lowers the effective sample size within a split but does not leak across splits. The imbalanced mixture takes the first 4K, 4K, 8K, 9K, 7K, and 4K training images of the same sources. The validation split is the only evaluation split of this study.

\subsection{Compute per sample}\label{app:compute}

Table~\ref{tab:compute} counts the agent passes that each regime needs per training sample and per evaluated sample. Every regime first runs the frozen DINOv3 encoder once per image. Solo, random, and K-means routing train one agent per sample. Learned central routing adds a router. Distributed delegation needs a forward pass of the requester, because its router reads the requester's queries, before the helper is trained. \methodName additionally trains the requester on its helped reconstruction. At evaluation time, delegation runs two agents in the distributed regimes (the requester for routing, the helper for reconstruction) and collaboration three, plus the transfer of the helper's keys and values in four layers. A four-agent population has four times the parameters of the solo model but the same number of active parameters per forward pass. A parameter- or FLOP-matched single model is not part of this study; the random-routing control shows only that four times the parameters do not help by themselves under the fixed sample budget.

\begin{table}[!htbp]
\centering
\caption{Agent passes per sample for a four-agent type-B population (fwd/bwd: forward and backward passes). Evaluation counts the forward passes of the delegation and collaboration operating points.}
\label{tab:compute}
\small
\setlength{\tabcolsep}{5pt}
\begin{tabular}{@{}l cc cc@{}}
\toprule
 & \multicolumn{2}{c}{Training, per sample} & \multicolumn{2}{c}{Evaluation, agent fwd} \\
\cmidrule(lr){2-3}\cmidrule(lr){4-5}
Regime & agent fwd & agent bwd & Deleg. & Collab. \\
\midrule
Solo & 1 & 1 & 1 & -- \\
Random, K-means & 1 & 1 & 1 & 2 \\
Learned central & 1 (+router) & 1 (+router) & 1 & 2 \\
Distr.\ delegation & 2 & 1 (+router) & 2 & 3 \\
\methodName & 3 & 2 (+router) & 2 & 3 \\
\bottomrule
\end{tabular}
\end{table}

\needspace{8\baselineskip}
\subsection{Method}\label{app:method}
\begin{algorithm}[h]
\caption{One \methodName training step for a population \( \{a_i, r_i\}_{i=1}^K \)}
\label{alg:disco}
\begin{algorithmic}
\STATE Sample a batch \( \{x_i = (\hat z_i, z_i)\}_{i=1}^b \) and a requester index \( q \sim \mathrm{Uniform}\{1,\dots,K\} \)
\STATE For each sample \( i \), draw a helper \( j \sim r_q(\hat z_i) \) with \( r_q^{(q)} \equiv 0 \) (no self-selection)
\STATE Helpers: compute \( l(x_i, a_j) \) and their keys/values \( K^l_j, V^l_j \) in communication layers
\STATE Requester: compute \( l_{\rm solo}(x_i, a_q) \) and \( l_{\rm helped}(x_i, a_q, a_j) \) using \( {\rm sg}[K^l_j], {\rm sg}[V^l_j] \)
\STATE Router reward \( R_i \gets {\rm sg}[\, l_{\rm solo} - l_{\rm helped}\,] \); update the requester's baseline \( \bar R_q \)
\STATE Update: helpers' LoRA with \( \mathcal{L}_{\rm rec} \); requester's LoRA with \( \mathcal{L}_{\rm helped} \); \( r_q \) with \( \mathcal{L}_{\rm router} + \lambda_{\rm KL}\mathcal{L}_{\rm KL} \)
\end{algorithmic}
\end{algorithm}

We use DINOv3 ViT-B/16~\citep{simeoni2025dinov3} as the frozen encoder, normalize its features before feeding them to the agents, and apply the cosine distance loss only on the masked part of the feature sequence.

During training the helper index is sampled from the distribution predicted by the requester's router, with logits divided by a temperature $\tau = 1.25$. To foster specialization, the requester is excluded from its own routing distribution: its probability is set to 0 and the remaining probabilities are renormalized. Each requester keeps its own exponential-moving-average baseline of improvements with momentum 0.95. The loss weights are $\lambda_{\rm rec} = 1.25$, $\lambda_{\rm router} = 1.0$, and $\lambda_{\rm helped} = 1.0$. All learned routers are regularized with a KL divergence between the marginal routing distribution and the uniform distribution,
\begin{align}
    {\cal L}_{\rm KL} = \sum_{j = 1}^K \bar r^{(j)} \log \bigl[K\cdot \bar r^{(j)}\bigr], \;{\rm where}\; \bar r^{(j)} = \frac{1}{b}\sum_{i = 1}^b r_q^{(j)}(\hat z_i)
\end{align}
is the marginal, over the current batch, of the routing distribution of the active router \( r_q \) (the central router, or the requester's router in the distributed regimes). It is added with weight $\lambda_{\rm KL} = 0.02$ in all learned-routing regimes. This is the standard load-balancing mechanism of MoE training. As the random-routing control shows, balance alone does not produce specialization; Section~\ref{sec:res_robust} and Appendix~\ref{app:seeds} ablate it.

\FloatBarrier
\subsection{Specialization dynamics of the five routing regimes}\label{app:dynamics}

\begin{figure}[!htbp]
\centering
\begin{minipage}[t]{0.575\linewidth}
    \centering
    \includegraphics[height=4.0cm]{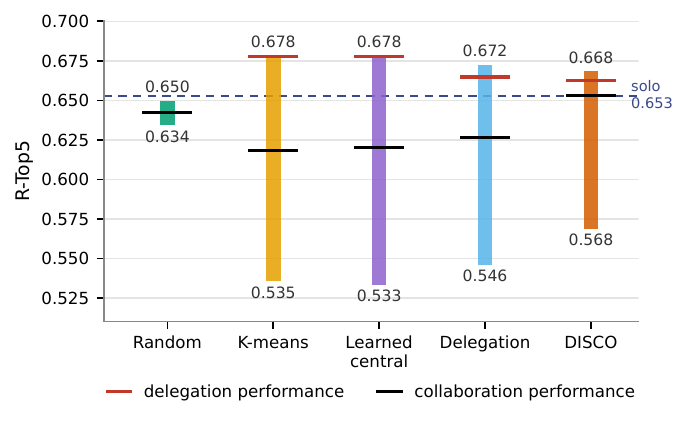}
    \captionof{figure}{Performance envelopes of the routing regimes (values of Table~\ref{tab:main}). Bars span R-Top5-Worst to R-Top5-Best; the red and black marks are the delegation and the collaboration performance.}
    \label{fig:routing_perf}
\end{minipage}
\hfill
\begin{minipage}[t]{0.40\linewidth}
    \centering
    \includegraphics[height=4.0cm]{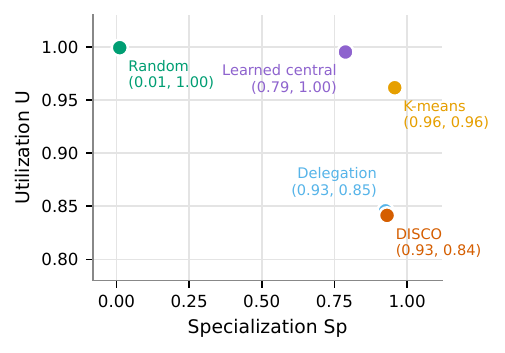}
    \captionof{figure}{Specialization \( \mathrm{Sp} \) versus utilization \( \mathrm{U} \) of the same populations. Random routing is balanced but not specialized; the decentralized regimes are as specialized as the K-means split, at somewhat lower balance.}
    \label{fig:routing_sp}
\end{minipage}
\end{figure}

\begin{figure}[!htbp]
\centering
\includegraphics[width=0.24\linewidth]{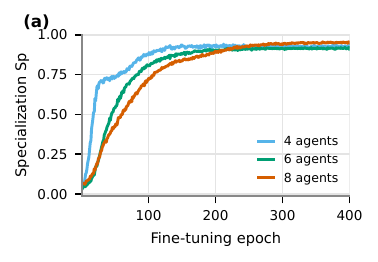}\hfill
\includegraphics[width=0.24\linewidth]{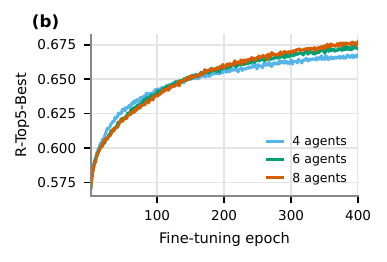}\hfill
\includegraphics[width=0.24\linewidth]{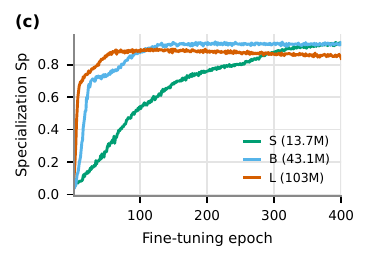}\hfill
\includegraphics[width=0.24\linewidth]{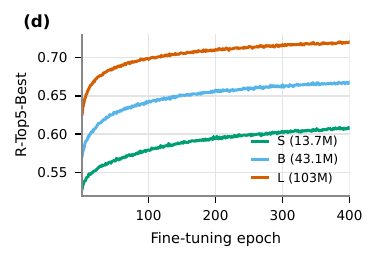}
\caption{Scaling of \methodName populations during training (training-time validation): Sp and R-Top5-Best. (a,b) Populations of 4, 6, and 8 type-B agents: larger populations specialize later and end higher. (c,d) Populations of four S, B, or L agents: more capable agents specialize earlier and perform better. Discussed in Section~\ref{sec:res_scaling}.}
\label{fig:scaling}
\end{figure}

\begin{figure}[!htbp]
    \centering
    \includegraphics[width=0.9\linewidth]{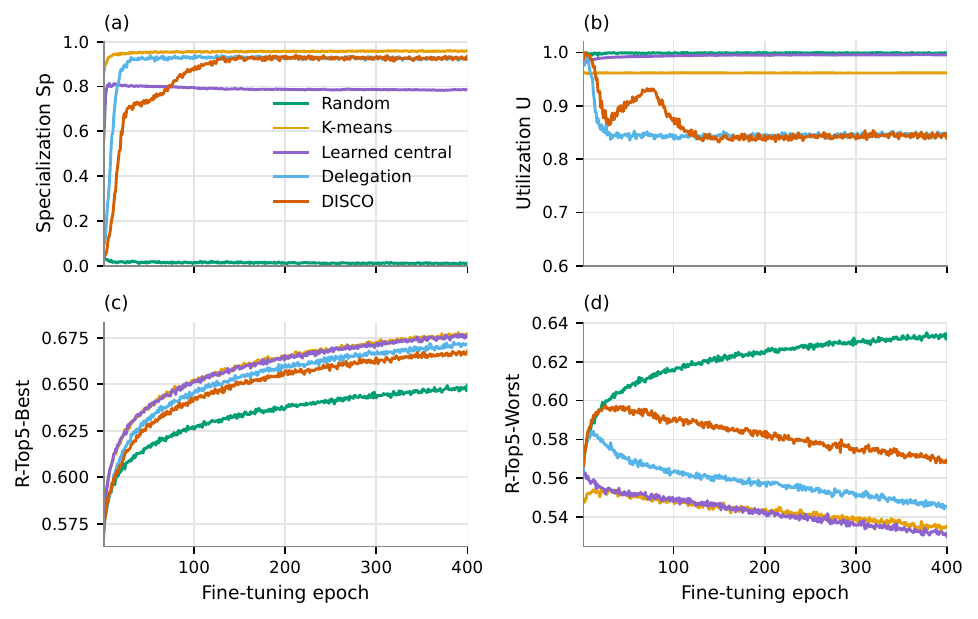}
    \caption{Specialization \( \mathrm{Sp} \) (a), utilization \( \mathrm{U} \) (b), R-Top5-Best (c) and R-Top5-Worst (d) during fine-tuning for the five routing regimes of Table~\ref{tab:main} (training-time validation, 4 type-B agents). K-means is specialized from the first epoch by construction; the learned central router settles at \( \mathrm{Sp}\approx0.79 \) within a few epochs; delegation passes \( \mathrm{Sp}=0.8 \) after 16 epochs and \methodName after 72. Utilization starts near 1 everywhere and stays there under random and learned central routing. It falls only in the two distributed regimes, where responsibility concentrates on three of the four agents, and in \methodName it rebounds for a few tens of epochs before settling at the level of delegation (\( \mathrm{U}\approx0.84 \)). Panels (c) and (d) show the price of specialization: Best rises together with Sp, while Worst falls for every specialized population and falls least for \methodName. Random routing stays at \( \mathrm{Sp}\approx0.01 \) and keeps the highest Worst.}
    \label{fig:ladder_curves}
\end{figure}

Figure~\ref{fig:ladder_curves} shows the training-time validation curves of all regimes. Both distributed regimes start from identical agents, so neither router has anything to learn from until the symmetry breaks. What differs is how quickly each reward reflects the break. Delegation is rewarded by the helper's standalone loss, which separates agents as soon as the random batches they are trained on differ, so its routers get a usable signal almost immediately. The reward of \methodName is a difference between two losses of the \emph{same} agent, the requester's reconstruction with and without the message, and that difference only becomes informative once the helpers have begun to differ \emph{and} the requester has learned to read their keys and values. It is therefore smaller and noisier early in training, which is what delays the onset of specialization (72 against 16 epochs to \( \mathrm{Sp}=0.8 \)). In the last 20 fine-tuning epochs the curves fluctuate with a standard deviation of 0.0005--0.0009 (R-Top5-Best), 0.0007--0.0018 (R-Top5-Worst), 0.001--0.005 (Sp), and below 0.004 (U) for every regime, which we use as the scale of the evaluation noise in Section~\ref{sec:experiments}. Table~\ref{tab:agreement} evaluates the regular checkpoints of the \methodName and delegation populations with the protocol of Appendix~\ref{app:eval_protocol}: the helped score of \methodName rises with the experts' sharpening, while that of delegation, whose agents never learned to read a helper, stays flat. Moreover, the routers agree on the expert for 98\% of the inputs from the earliest saved checkpoint on, so agreement is established long before specialization has finished deepening.

\begin{table}[!htbp]
\centering
\caption{Routing agreement and specialization along training for the 4-agent \methodName and delegation populations (checkpoints every 50 fine-tuning epochs from epoch 150, one draw of crops and masks, 13,140 validation images). Agree: fraction of inputs on which every non-expert requester's most probable helper is the expert. Agreement is complete by epoch 150 and stays at 98\%, while the populations keep specializing (Worst falls) and, in \methodName only, the helped requester keeps improving (Collab.\ and Worst+h.\ rise). Sp and U over the same checkpoints are in Figure~\ref{fig:ladder_curves}.}
\label{tab:agreement}
\small
\setlength{\tabcolsep}{4pt}
\begin{tabular}{@{}l l ccccc@{}}
\toprule
Epoch & Regime & Agree & Best & Worst & Collab. & Worst+h. \\
\midrule
150 & \methodName & \best{0.985} & 0.651 & \best{0.587} & \best{0.638} & \best{0.629} \\
 & Delegation & 0.981 & \best{0.654} & 0.562 & 0.623 & 0.604 \\
\addlinespace[1pt]
200 & \methodName & \best{0.982} & 0.657 & \best{0.584} & \best{0.642} & \best{0.634} \\
 & Delegation & 0.971 & \best{0.660} & 0.558 & 0.625 & 0.603 \\
\addlinespace[1pt]
250 & \methodName & \best{0.982} & 0.661 & \best{0.579} & \best{0.646} & \best{0.637} \\
 & Delegation & \best{0.982} & \best{0.665} & 0.554 & 0.626 & 0.601 \\
\addlinespace[1pt]
300 & \methodName & \best{0.981} & 0.665 & \best{0.576} & \best{0.649} & \best{0.641} \\
 & Delegation & 0.980 & \best{0.667} & 0.552 & 0.626 & 0.597 \\
\addlinespace[1pt]
350 & \methodName & \best{0.980} & 0.666 & \best{0.572} & \best{0.651} & \best{0.642} \\
 & Delegation & 0.979 & \best{0.669} & 0.548 & 0.626 & 0.595 \\
\addlinespace[1pt]
400 & \methodName & \best{0.981} & 0.668 & \best{0.568} & \best{0.653} & \best{0.644} \\
 & Delegation & 0.980 & \best{0.672} & 0.546 & 0.626 & 0.593 \\
\bottomrule
\end{tabular}

\end{table}

\begin{table}[!htbp]
\centering
\caption{Final checkpoints of the scaling study evaluated with the protocol of Appendix~\ref{app:eval_protocol} (mixture validation set, three draws). With more agents the oracle Best of \methodName rises and stays 0.007--0.010 below a K-means split of the same size, whose collaboration score is 0.03--0.05 lower. The helped requester of the 8-agent population exceeds the solo model (0.655 vs.\ 0.653). Agreement decreases with population size (98\%, 98\%, 88\%) because each router is trained on fewer batches. Capacity moves every score in the same direction: from four S to four B to four L agents, Best rises 0.607, 0.668, 0.721, Collab.\ 0.591, 0.653, 0.709 and Worst+h.\ 0.580, 0.644, 0.704, and the help closes more of the gap to the expert as the agents grow (Worst+h.\ is 0.027 below Best for S, 0.024 for B and 0.017 for L). Specialization stays high for S and B (\( \mathrm{Sp}\approx0.93 \)) and is lower for L (0.84) at a higher utilization (0.91). Each population is initialized from a pretrained model of its own size.}
\label{app:scaling_table}
\small
\setlength{\tabcolsep}{4pt}
\begin{tabular}{@{}l cccccccc@{}}
\toprule
Population & Sp & U & Best & Worst & Deleg. & Collab. & Worst+h. & Agree \\
\midrule
\methodName, 4 B agents & \second{0.931} & 0.841 & 0.668 & \second{0.568} & 0.662 & \second{0.653} & 0.644 & \best{0.983} \\
\methodName, 6 B agents & 0.914 & \best{0.998} & \second{0.674} & \best{0.579} & \second{0.666} & 0.652 & \second{0.645} & \second{0.982} \\
\methodName, 8 B agents & \best{0.953} & \second{0.936} & \best{0.678} & 0.543 & \best{0.673} & \best{0.655} & \best{0.647} & 0.881 \\
\midrule
K-means, 4 B agents & \best{0.957} & \best{0.962} & 0.678 & \best{0.535} & 0.678 & \best{0.618} & \best{0.582} & -- \\
K-means, 6 B agents & 0.887 & \second{0.935} & \second{0.681} & \second{0.532} & \second{0.679} & \second{0.607} & \second{0.573} & -- \\
K-means, 8 B agents & \second{0.928} & 0.912 & \best{0.685} & 0.530 & \best{0.683} & 0.606 & 0.572 & -- \\
\midrule
\methodName, 4 S agents & \best{0.926} & 0.843 & 0.607 & 0.528 & 0.602 & 0.591 & 0.580 & \best{0.979} \\
\methodName, 4 L agents & 0.841 & \best{0.909} & \best{0.721} & \best{0.612} & \best{0.715} & \best{0.709} & \best{0.704} & 0.939 \\
\bottomrule
\end{tabular}

\end{table}

\FloatBarrier
\subsection{Independent fine-tuning seeds and protocol ablations}\label{app:seeds}

Table~\ref{tab:seeds} lists all replication and ablation runs. Every run uses the configuration of its regime in Section~\ref{sec:regimes} with a new seed, which changes the LoRA and router initialization, the data order, the requester and helper draws, and the masks; the same seed gives the same data order in every regime, so rows with equal seeds are paired. Runs last 100 fine-tuning epochs, a quarter of the budget of Table~\ref{tab:main}, and the final checkpoint is evaluated with the protocol of Appendix~\ref{app:eval_protocol} on one draw of crops and masks.

\emph{Seeds.} Training-time validation places the onset of specialization (first validation with \( \mathrm{Sp}\ge0.8 \)) at fine-tuning epoch 28, 28, and 32 for \methodName and 16, 28, and 16 for delegation. Two \methodName populations and one delegation population use all four agents (expert shares 17--33\%); the others leave one agent with 2\% (\methodName, seed 2) or 9--10\% (delegation, seeds 1--2) of the inputs. Whether an agent is left idle is thus a property of the run, not of the regime. The same-requester interventions of Table~\ref{tab:interventions} hold for every seed: a \methodName requester scores 0.614--0.620 alone, 0.631--0.632 with its selected helper, 0.622--0.624 with a random other agent, 0.600--0.606 with zero keys and values, and 0.395--0.398 with a shuffled message (delegation: 0.602--0.608, 0.621--0.623, 0.607--0.611, 0.592--0.598, 0.404--0.405).

\emph{Ladder.} The ordering of Section~\ref{sec:res_central} replicates at this budget: three solo seeds score 0.633--0.634, two random-routing populations 0.628 (\( \mathrm{Sp}=0.02 \), \( \mathrm{U}=1.00 \)), the distributed regimes 0.645--0.646, and two learned central routers 0.651--0.652 (\( \mathrm{Sp}=0.70 \)--\( 0.81 \), \( \mathrm{U}=1.00 \)). The helped \methodName requester (0.631--0.632) is 0.002--0.003 below the solo model at this budget (on par with it at the full budget, Table~\ref{tab:main}). On ImageNet100 the solo seeds score 0.654--0.655, \methodName's collaboration 0.657--0.658 and its delegation 0.656, delegation's collaboration 0.655--0.656, and learned central routing 0.651--0.654 with communication against 0.654 without, so \methodName remains the only regime whose collaboration exceeds both the solo model and its own delegation.

\emph{Reward versus channel.} The cell ``delegation reward + communication training'' trains the requester on its helped loss but rewards its router with the helper's loss, as in delegation; ``random routing + communication training'' draws the helper uniformly among the other agents and trains no router (its Deleg.\ column is the score of an average agent and its Collab.\ column the random-helper collaboration). The first reproduces the helped scores of \methodName (collaboration 0.631, Worst+help 0.624) together with delegation's specialization and routing (\( \mathrm{Sp}=0.86 \), \( \mathrm{U}=0.93 \), agreement 82\%, onset after 20 epochs, least-used agent 8\%). The second never breaks the symmetry (\( \mathrm{Sp}=0.00 \), \( \mathrm{U}=1.00 \), expert shares 24--26\%); its agents reach 0.630 as experts and 0.622 as worst agents, and a random helper's message is worth 0.001 to a requester (0.627 vs.\ 0.626).

\emph{Protocol choices.} With \( \lambda_{\rm KL}=0 \) the \methodName population collapses onto one agent within the first four fine-tuning epochs and stays there: the routers of three agents send every input to agent~2 and agent~2's own router sends everything to agent~1, so agent~2 is trained on three quarters of the traffic and is the expert on 90\% of the inputs (expert shares 9\%, 90\%, 1\%, 0\%; \( \mathrm{Sp}=0.08 \), \( \mathrm{U}=0.25 \)). Its Best (0.633) equals the solo model of the same budget and its Worst (0.566) that of a specialized population; a helped requester still profits from the generalist's message (0.628 vs.\ 0.599 alone, 0.588 with zero and 0.405 with shuffled keys and values), but nothing is divided. The KL term is thus a precondition for the division of labor under local routing, as load balancing is for MoEs; the random-routing control shows that it is not sufficient. With self-selection allowed (the requester's own index is not removed from its routing distribution), specialization still emerges, later and weaker (\( \mathrm{Sp}=0.81 \), \( \mathrm{U}=0.99 \); Sp passes 0.8 after 72 epochs against 28--32 with exclusion): agents keep the aerial images and faces to single experts and split the two robotics domains and the driving images between two agents. The routers select their own agent on 24\% of the inputs, on 91\% of those on which their agent is the expert and on 2\% of the others, and non-expert requesters select the expert on 93\% of the inputs, so the exclusion sharpens and accelerates the division of labor but is not what creates it. The helped scores equal those of \methodName (collaboration 0.633, Worst+help 0.625; a requester alone 0.604, with zero keys and values 0.591, shuffled 0.391).

\major{\begin{table}[!htbp]
\centering
\caption{Replication and ablations at a quarter of the training budget (100 fine-tuning epochs, seeded protocol of Appendix~\ref{app:eval_protocol}, one draw of crops and masks). Seeds 1--3 change the LoRA and router initialization, the data order, the requester and helper draws, and the masks; the same seed gives the same data order in every regime. Mean \( \pm \) standard deviation over seeds where several seeds exist.}
\label{tab:seeds}
\footnotesize
\setlength{\tabcolsep}{3pt}
\begin{adjustbox}{max width=\linewidth}
\begin{tabular}{@{}l c cccccccc@{}}
\toprule
Regime & Seed & Sp & U & Best & Worst & Deleg. & Collab. & Worst+help & Agree \\
\midrule
Solo & 1 & -- & -- & 0.633 & 0.633 & -- & -- & -- & -- \\
Solo & 2 & -- & -- & 0.634 & 0.634 & -- & -- & -- & -- \\
Solo & 3 & -- & -- & 0.634 & 0.634 & -- & -- & -- & -- \\
\quad mean $\pm$ sd & & -- & -- & 0.634\,{\scriptsize$\pm$.001} & 0.634\,{\scriptsize$\pm$.001} & -- & -- & -- & -- \\
Random & 1 & 0.018 & 0.997 & 0.628 & 0.617 & 0.622 & 0.623 & 0.619 & -- \\
Random & 2 & 0.018 & 0.999 & 0.628 & 0.617 & 0.622 & 0.623 & 0.619 & -- \\
\quad mean $\pm$ sd & & 0.018\,{\scriptsize$\pm$.000} & 0.998\,{\scriptsize$\pm$.002} & 0.628\,{\scriptsize$\pm$.000} & 0.617\,{\scriptsize$\pm$.000} & 0.622\,{\scriptsize$\pm$.000} & 0.623\,{\scriptsize$\pm$.000} & 0.619\,{\scriptsize$\pm$.000} & -- \\
Learned central & 1 & 0.808 & 0.999 & 0.652 & 0.557 & 0.652 & 0.615 & 0.600 & -- \\
Learned central & 2 & 0.700 & 0.998 & 0.651 & 0.550 & 0.651 & 0.615 & 0.598 & -- \\
\quad mean $\pm$ sd & & 0.754\,{\scriptsize$\pm$.076} & 0.999\,{\scriptsize$\pm$.001} & 0.652\,{\scriptsize$\pm$.001} & 0.553\,{\scriptsize$\pm$.005} & 0.651\,{\scriptsize$\pm$.001} & 0.615\,{\scriptsize$\pm$.000} & 0.599\,{\scriptsize$\pm$.002} & -- \\
Distr.\ delegation & 1 & 0.863 & 0.943 & 0.646 & 0.565 & 0.640 & 0.622 & 0.607 & 0.820 \\
Distr.\ delegation & 2 & 0.855 & 0.932 & 0.646 & 0.571 & 0.640 & 0.623 & 0.608 & 0.818 \\
Distr.\ delegation & 3 & 0.950 & 0.964 & 0.646 & 0.566 & 0.640 & 0.621 & 0.607 & 0.985 \\
\quad mean $\pm$ sd & & 0.889\,{\scriptsize$\pm$.053} & 0.946\,{\scriptsize$\pm$.016} & 0.646\,{\scriptsize$\pm$.000} & 0.568\,{\scriptsize$\pm$.003} & 0.640\,{\scriptsize$\pm$.000} & 0.622\,{\scriptsize$\pm$.001} & 0.607\,{\scriptsize$\pm$.001} & 0.874\,{\scriptsize$\pm$.096} \\
\methodName & 1 & 0.948 & 0.963 & 0.644 & 0.601 & 0.639 & 0.631 & 0.624 & 0.986 \\
\methodName & 2 & 0.919 & 0.846 & 0.645 & 0.585 & 0.640 & 0.632 & 0.625 & 0.980 \\
\methodName & 3 & 0.942 & 0.963 & 0.645 & 0.597 & 0.639 & 0.631 & 0.625 & 0.986 \\
\quad mean $\pm$ sd & & 0.936\,{\scriptsize$\pm$.015} & 0.924\,{\scriptsize$\pm$.068} & 0.645\,{\scriptsize$\pm$.000} & 0.594\,{\scriptsize$\pm$.009} & 0.639\,{\scriptsize$\pm$.001} & 0.631\,{\scriptsize$\pm$.001} & 0.625\,{\scriptsize$\pm$.001} & 0.984\,{\scriptsize$\pm$.004} \\
Delegation reward + comm.\ training & 1 & 0.855 & 0.926 & 0.645 & 0.586 & 0.640 & 0.631 & 0.624 & 0.818 \\
Random routing + comm.\ training & 1 & 0.002 & 1.000 & 0.630 & 0.622 & 0.626 & 0.627 & 0.625 & 0.004 \\
\methodName, \( \lambda_{\rm KL}=0 \) & 1 & 0.084 & 0.255 & 0.633 & 0.566 & 0.631 & 0.628 & 0.625 & 0.905 \\
\methodName, self-selection allowed & 1 & 0.812 & 0.995 & 0.647 & 0.572 & 0.646 & 0.633 & 0.625 & 0.915 \\
\bottomrule
\end{tabular}
\end{adjustbox}
\end{table}
}

\FloatBarrier
\subsection{Which domains each agent takes, and where help pays off}\label{app:assignment}

\begin{figure}[!htbp]
    \centering
    \includegraphics[width=0.98\linewidth]{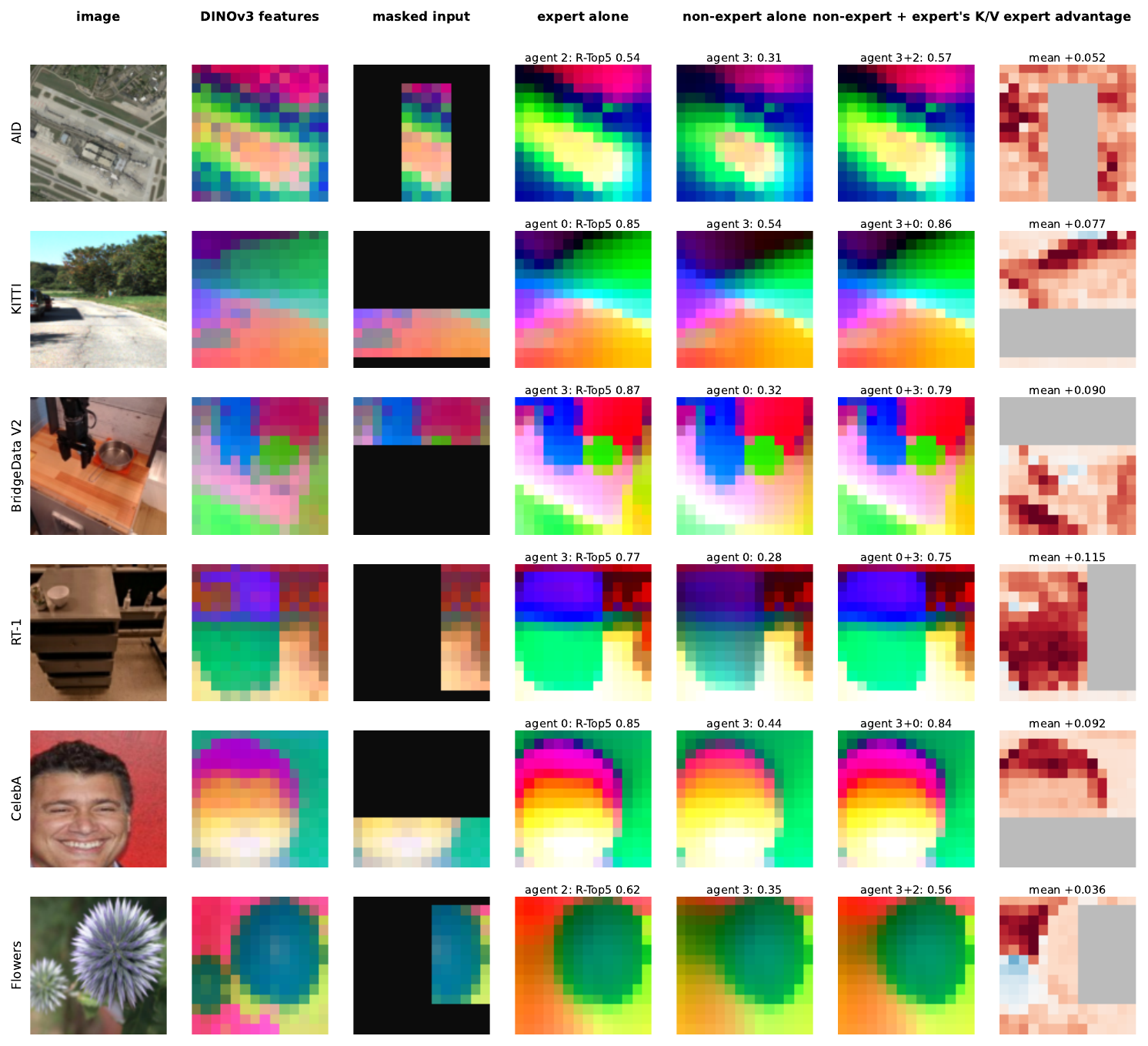}
    \caption{What the expert knows and the non-expert does not. One validation image per domain, 4-agent \methodName population, same crops and masks as in Table~\ref{tab:main}. DINOv3 patch features are projected to RGB by a PCA fitted on the target features of each image. The masked input shows the 30\% visible window. Columns 4--6: reconstructions of the expert (lowest-loss agent) alone, of the highest-loss agent alone, and of the same non-expert reading the expert's keys and values, with their R-Top5. Last column: per-patch reconstruction error (cosine distance to the DINOv3 target) of the non-expert minus that of the expert, on the hidden patches (red: the expert is better; visible window grey; mean in the title). The non-expert misses whole structures, e.g., the pan and gripper in BridgeData~V2, the road--vegetation boundary in KITTI, the hairline in CelebA, which the expert reconstructs and the help restores. Examples are those with the largest help gain per domain.}
    \label{fig:reconstructions}
\end{figure}

\begin{figure}[!htbp]
    \centering
    \includegraphics[width=\linewidth]{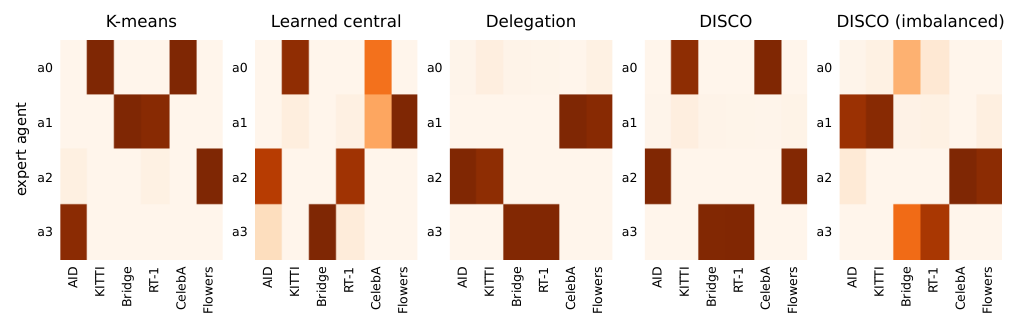}
    \caption{Fraction of each domain's validation images whose expert (lowest standalone loss) is agent \( a_i \), for the specialized regimes of Table~\ref{tab:main} and the imbalanced mixture. Delegation and \methodName both split the six domains into three pairs and leave one agent almost idle, which is what lowers their utilization to \( \mathrm{U}\approx0.84 \), but the pairs differ: delegation groups aerial with driving and faces with flowers, \methodName groups aerial with flowers and driving with faces, and only the two robotics datasets are kept together by both; the learned central router splits the face images across two agents; K-means separates aerial images from flowers. On the imbalanced mixture the two robotics datasets (8K and 9K training images) are split between two agents instead.}
    \label{fig:assignment}
\end{figure}

\begin{figure}[!htbp]
    \centering
    \begin{minipage}[b]{0.34\linewidth}
        \centering
        \includegraphics[width=\linewidth]{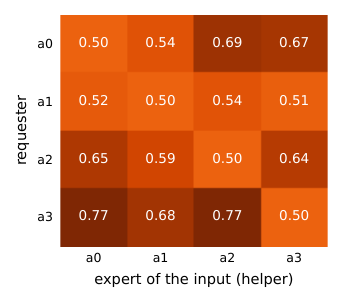}
    \end{minipage}
    \hfill
    \begin{minipage}[b]{0.62\linewidth}
        \centering
        \includegraphics[width=\linewidth]{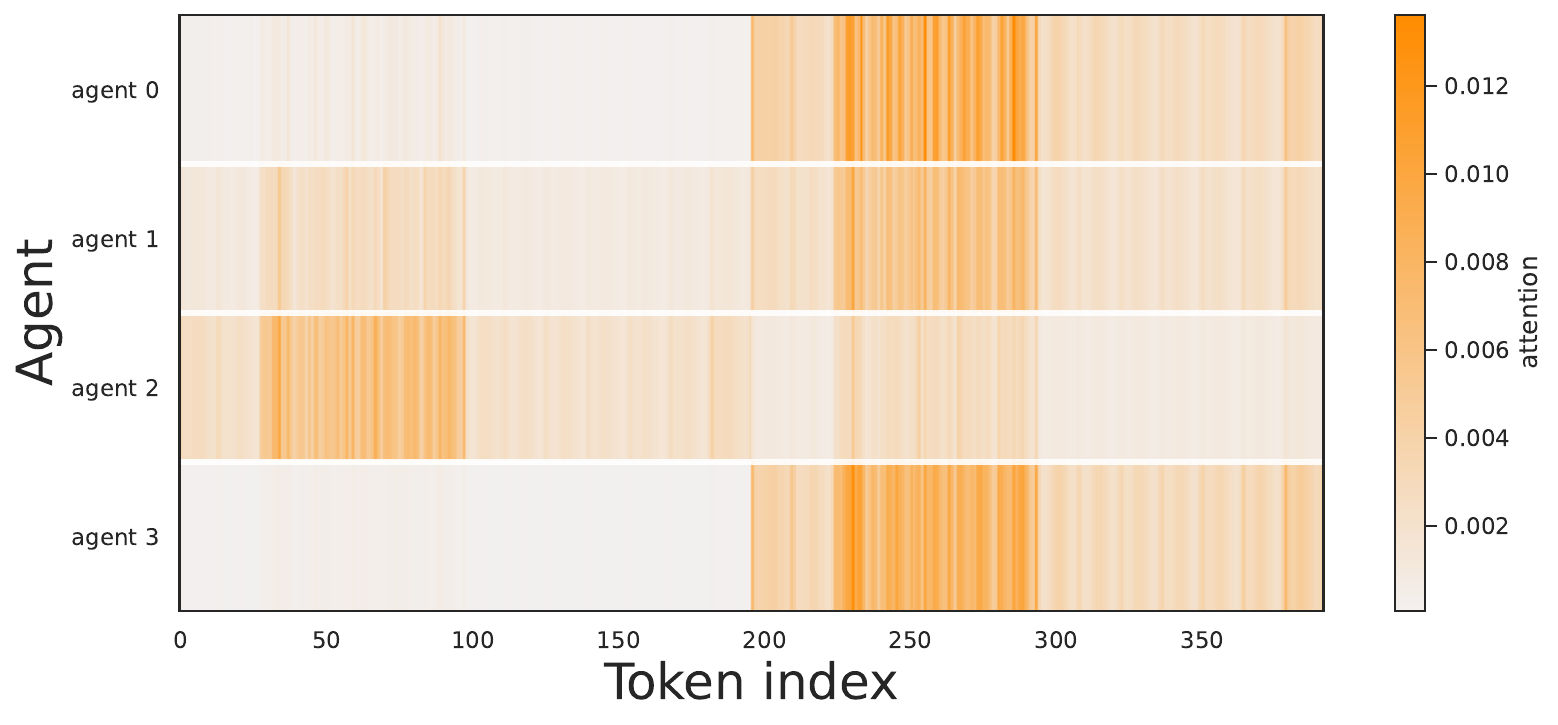}
    \end{minipage}
    \caption{Where the requesters look. \textbf{Left:} attention mass that requester \( a_i \) (rows) places on the helper's keys and values in the communication layers of the 4-agent \methodName population, on inputs whose expert is agent \( a_j \) (columns), averaged over heads, queries, layers, and the validation set. When requester and helper coincide (diagonal) the two copies of the same keys receive half of the mass each, so 0.5 is the reference. The three specialists place 0.64--0.77 of their attention on the expert's tokens, while the almost idle agent \( a_1 \) (Figure~\ref{fig:assignment}) uses helpers little (0.51--0.54). \textbf{Right:} per-token attention of every agent in the last communication layer, for one input whose expert is agent~2 (mean over heads and queries; token indices 0--195 are the requester's own keys, 196--391 the helper's). The expert attends to its own keys; the other agents attend mostly to the expert's.}
    \label{fig:attention_matrix}
\end{figure}

In this section, we look at what the agents divide and at what the help actually restores. Figure~\ref{fig:reconstructions} makes the division concrete: on one validation image per domain, the expert reconstructs structures that the highest-loss agent misses entirely, and the same non-expert recovers most of them once it reads the expert's keys and values, which is the qualitative counterpart of the Worst+help column of Table~\ref{tab:main}. Figure~\ref{fig:assignment} then shows which domains each agent takes in every specialized population, Table~\ref{tab:per_domain} breaks the scores down by domain, and Figure~\ref{fig:attention_matrix} shows how much of their attention requesters spend on the helper's tokens, over the validation set (left) and for a single input (right).

\begin{table}[!htbp]
\centering
\caption{Per-domain R-Top5 on the mixture validation set (2,190 images per domain). Best is the expert's score, Worst the highest-loss agent's, Collab.\ the score of a random requester helped by its selected helper, Worst+help that of the worst agent helped. Solo scores show that the domains differ widely in difficulty. The collaboration gain of \methodName over delegation is largest on faces, followed by the two robotics domains, and is absent on aerial images, the domain on which delegation leaves its idle agent.}
\label{tab:per_domain}
\small
\setlength{\tabcolsep}{4pt}
\begin{tabular}{@{}l cccccc@{}}
\toprule
Regime / measure & AID & KITTI & Bridge & RT-1 & CelebA & Flowers \\
\midrule
Solo & 0.429 & 0.741 & 0.694 & 0.724 & 0.788 & 0.540 \\
K-means, Best (expert) & \best{0.462} & 0.757 & \second{0.733} & \second{0.750} & \best{0.805} & \best{0.563} \\
Learned central, Best (expert) & 0.444 & \best{0.762} & \best{0.749} & \best{0.754} & \second{0.802} & \second{0.558} \\
Delegation, Best (expert) & \second{0.448} & \second{0.759} & 0.726 & 0.747 & \second{0.802} & 0.553 \\
\methodName, Best (expert) & 0.442 & 0.754 & 0.719 & 0.742 & 0.800 & 0.554 \\
\midrule
Delegation, Worst & \best{0.407} & 0.664 & 0.515 & 0.541 & 0.666 & 0.481 \\
\methodName, Worst & 0.390 & \best{0.687} & \best{0.541} & \best{0.590} & \best{0.697} & \best{0.506} \\
\midrule
Delegation, Collab. & \best{0.433} & 0.729 & 0.653 & 0.683 & 0.735 & 0.524 \\
\methodName, Collab. & 0.430 & \best{0.743} & \best{0.692} & \best{0.722} & \best{0.787} & \best{0.543} \\
\midrule
Delegation, Worst+help & \best{0.437} & 0.707 & 0.607 & 0.639 & 0.663 & 0.502 \\
\methodName, Worst+help & 0.424 & \best{0.738} & \best{0.678} & \best{0.711} & \best{0.777} & \best{0.536} \\
\bottomrule
\end{tabular}

\end{table}

\FloatBarrier
\subsection{Evaluation protocol}\label{app:eval_protocol}

Sp and U are computed on the validation sets of the six domains (2,190 images each, 13,140 in total) from the standalone losses of the agents on the same masked inputs. The routers play no role in these two metrics. R-Top5-Best and R-Top5-Worst are the expert envelope over agents (Section~\ref{sec:setting}). The \emph{delegation} performance evaluates the population with its own router: each validation sample is reconstructed by the router's most probable agent (for a random requester in the distributed regimes, whose own index is excluded). For K-means and random routing the fixed assignment of the sample is used. The \emph{collaboration} performance draws a random requester for each sample and lets it reconstruct with the keys and values of the selected helper. For the held-out evaluation (Section~\ref{sec:res_disco}), the same protocol is applied to the ImageNet100 validation set, which none of the populations saw during specialization. The routing-agreement statistic (Section~\ref{sec:res_routing}) counts validation inputs for which every non-expert requester's most likely helper is the expert \( e(x) \).
\paragraph{Metric.} Two model-free predictors calibrate R-Top5 on the same inputs: copying the feature of the spatially nearest visible patch into every hidden position scores 0.120, predicting the mean of the visible features scores 0.001, and chance is \( 5/196 = 0.026 \); the pretrained agent before specialization scores 0.563 and the solo model 0.653 (Table~\ref{tab:main}). The expert is selected by loss and scored by R-Top5. The lowest-loss agent is also the agent with the highest R-Top5 on 71\% (\methodName), 73\% (delegation), 84\% (learned central), and 87\% (K-means) of the inputs, but only on 41\% under random routing, whose agents are nearly interchangeable. Averaging instead the per-image maximum of R-Top5 over agents gives 0.673, 0.678, 0.683, 0.682, and 0.660 for these five regimes, 0.004--0.010 above the loss-selected Best of Table~\ref{tab:main}, and the per-image minimum gives 0.560, 0.524, 0.511, 0.511, and 0.624, 0.008--0.024 below the loss-selected Worst. The loss-selected envelope is thus slightly narrower, and the ordering of the specialized regimes is the same under both. The random population is the exception: its R-Top5 maximum (0.660) would exceed the solo model (0.653), because maximizing the reported metric over four nearly interchangeable agents selects evaluation noise. This is why the expert is defined by an independent criterion, the loss, and why we do not report the metric's own extrema as the envelope.
\paragraph{Message interventions.} Table~\ref{tab:interventions} uses the same cached inputs and the same requester--helper pairs as Table~\ref{tab:main}; the shuffled message is the selected helper's keys and values for the next image of the evaluation batch, and the zero message has the same token count.
\paragraph{Seeded protocol and intervals.} Validation uses the training-time augmentation (random resized crop and flip) and a random 70\% mask per image, so an evaluation depends on the draw. All numbers in Section~\ref{sec:experiments} come from one protocol: three independent draws of crops and masks are fixed per image and shared by all regimes, every agent is run on every image, and every (requester, helper) pair is evaluated, so all comparisons are paired. Point estimates are means over the three draws. Three quantities describe how much a number can move, and they are not interchangeable.

\emph{(i) Draw-to-draw spread.} Re-drawing the crops and masks and repeating the whole evaluation changes a point estimate on the mixture by at most 0.0005 for Best, Deleg., Collab.\ and Worst+help, by 0.0012 for Worst, and by 0.005 for Sp. On ImageNet100 the spread of the R-Top5 quantities is at most 0.002.

\emph{(ii) Interval on a single entry.} Uncertainty from the finite validation set is a paired bootstrap over images (10,000 resamples, an image carrying its three draws). It asks how much one cell of Table~\ref{tab:main} would move if the 13,140 images were replaced by another sample of the same size. Because the domains differ widely in difficulty (Table~\ref{tab:per_domain}), that is the largest of the three: across all regimes and all R-Top5 columns the 95\% half-width is 0.002--0.003, and for Sp it reaches 0.007.

\emph{(iii) Interval on a difference between two regimes.} The same images, crops and masks are used for every regime, so a hard image lowers both scores at once and cancels out of the paired difference. Over the comparisons listed below, the 95\% half-width is 0.0002--0.0004 for Best and Deleg., up to 0.0007 for Collab., and up to 0.0015 for Worst and Worst+help, whose per-image values are more dispersed; on ImageNet100 the corresponding half-widths are 0.0005--0.0012. For Sp, which is a single population-level statistic rather than a per-image mean, the half-width on a difference is at most 0.005.

This is why the comparisons in Section~\ref{sec:experiments} are stated as paired differences: a gap of 0.02--0.03 between two regimes is resolved by an interval an order of magnitude narrower, even though the individual entries it is computed from carry a spread of the same order as the gap. The intervals below are exactly these paired differences, which is why they are much tighter than the entries in Table~\ref{tab:main}. Key differences: random \( - \) solo Best \( -0.003 \) \( [-0.003,-0.003] \); learned central \( - \) solo Best \( +0.026 \) \( [+0.025,+0.026] \); \methodName \( - \) delegation: Best \( -0.004 \) \( [-0.004,-0.004] \), Worst \( +0.023 \) \( [+0.022,+0.023] \), Collab.\ \( +0.026 \) \( [+0.026,+0.027] \), Worst+help \( +0.051 \) \( [+0.050,+0.052] \), Sp \( +0.005 \) \( [+0.000,+0.010] \); \methodName Collab.\ \( - \) solo \( +0.0002 \) \( [0.0000,+0.0004] \).

\section{Limitations}\label{app:limitations}

This study is a controlled investigation of emergent specialization, not a training recipe, and its evidence has clear boundaries.

\paragraph{Task and modality.} All experiments use masked reconstruction of frozen DINOv3 features. We argued in Section~\ref{sec:setting} why this is a reasonable proxy for predictive pretraining, but transfer of the findings to raw-pixel generation, video prediction, language modeling, multimodal learning, or downstream task optimization has not been tested. The key--value communication protocol itself is task-agnostic and applies to any transformer, so the protocol is not the obstacle; the cost of running population-level sweeps in those settings is.

\paragraph{Scale.} Populations have at most eight agents of at most 103M parameters, trained on 36K images. The uniform choice of the requester slows specialization as the population grows (Section~\ref{sec:res_scaling}). Much larger populations would need a different scheduling of who routes, and the interaction of specialization with web-scale data remains open.

\paragraph{Protocol choices.} Communication carries the keys and values of a single helper per sample and is deliberately gradient-free. We did not compare against a differentiable channel, nor against multi-helper or hierarchical communication. A differentiable channel would answer a different question: this study asks whether specialization and usable expertise can emerge when no gradient crosses between agents, and our results show that they can. Domain identity is used as the latent factor for evaluation. Specialization along factors that do not align with dataset boundaries would require other probes.

\section{Impact Statement}\label{app:impact_statement}

This work is foundational research on modular and collaborative neural systems. Its main potential benefit is to improve the efficiency, adaptability, and interpretability of machine learning models by encouraging populations of agents to divide labor across heterogeneous data. Such systems could eventually reduce the need for monolithic models by enabling more targeted use of capacity and by making it easier to analyze which components are responsible for different parts of a distribution.

At the same time, more adaptive multi-agent systems may inherit or amplify risks associated with the domains in which they are deployed. If used in high-stakes applications, emergent specialization could produce uneven performance across domains or user groups, especially when the latent structure of the data reflects sensitive or imbalanced attributes. We therefore view careful measurement of specialization, utilization, robustness, and fairness as important prerequisites before applying such methods outside controlled research settings. The present work does not introduce new datasets or deployment-ready models, and we do not anticipate direct societal harms from the experiments themselves.

\end{document}